\documentclass[lettersize,journal]{IEEEtran}
\usepackage{amsmath,amsfonts}
\usepackage{algorithmic}
\usepackage{algorithm}
\usepackage{array}
\usepackage[caption=false,font=normalsize,labelfont=sf,textfont=sf]{subfig}
\usepackage{textcomp}
\usepackage{stfloats}
\usepackage{url}
\usepackage{verbatim}
\usepackage{graphicx}

\usepackage{cite}
\usepackage{caption}
\usepackage{makecell}
\usepackage{contour}
\usepackage{wrapfig}
\usepackage{graphicx}%
\usepackage{multirow}%
\usepackage{booktabs}
\usepackage{adjustbox}
\usepackage{tcolorbox}
\usepackage{colortbl}
\PassOptionsToPackage{table}{xcolor}
\usepackage{booktabs}  %
\usepackage{adjustbox} %
\usepackage{colortbl}  %
\usepackage[table]{xcolor} %
\usepackage{hyperref}

\makeatletter
  \newcommand\figcaption{\def\@captype{figure}\caption}
  \newcommand\tabcaption{\def\@captype{table}\caption}
  
\makeatother
\definecolor{green(munsell)}{rgb}{0.0, 0.66, 0.47}
\definecolor{beaublue}{rgb}{0.74, 0.83, 0.9}
\definecolor{bluegray}{rgb}{0.4, 0.6, 0.8}
\definecolor{airforceblue}{rgb}{0.35, 0.5, 0.72}
\makeatletter
\renewcommand{\paragraph}{%
  \@startsection{paragraph}{4}%
  {\z@}{0.5em}{-0.0em}%
  {\normalfont\normalsize\bfseries}%
}
\makeatother

\hypersetup{
    colorlinks=true,   %
    citecolor=green(munsell),    %
    linkcolor=black,   %
    urlcolor=green(munsell)      %
}

\newcommand{\wj}[1]{{\color{black}#1}}

\begin{document}

\title{Toward a Holistic Evaluation of Robustness in CLIP Models}

\author{Weijie Tu, Weijian Deng, Tom Gedeon 
\thanks{W. Tu and W. Deng are with the School of Computing, The Australian National University, Canberra, ACT 0200, Australia. 
T. Gedeon is affiliated with The Australian National University, the University of Óbuda, and Curtin University. W. Deng is corresponding author.
E-mail: \{firstname.lastname\}@anu.edu.au}%
}

\markboth{Journal of \LaTeX\ Class Files,~Vol.~14, No.~8, August~2015}%
{Shell \MakeLowercase{\textit{et al.}}: Bare Advanced Demo of IEEEtran.cls for IEEE Computer Society Journals}

\maketitle

\begin{abstract}
Contrastive Language-Image Pre-training (CLIP) models have shown significant potential, particularly in zero-shot classification across diverse distribution shifts. Building on existing evaluations of overall classification robustness, this work aims to provide a more comprehensive assessment of CLIP by introducing several new perspectives. First, we investigate their robustness to variations in specific visual factors. Second, we assess two critical safety objectives—confidence uncertainty and out-of-distribution detection—beyond mere classification accuracy. Third, we evaluate the finesse with which CLIP models bridge the image and text modalities. Fourth, we extend our examination to 3D awareness in CLIP models, moving beyond traditional 2D image understanding. Finally, we explore the interaction between vision and language encoders within modern large multimodal models (LMMs) that utilize CLIP as the visual backbone, focusing on how this interaction impacts classification robustness.
In each aspect, we consider the impact of six factors on CLIP models: model architecture, training distribution, training set size, fine-tuning, contrastive loss, and test-time prompts.
Our study uncovers several previously unknown insights into CLIP. For instance, the architecture of the visual encoder in CLIP plays a significant role in their robustness against 3D corruption. CLIP models tend to exhibit a bias towards shape when making predictions. Moreover, this bias tends to diminish after fine-tuning on ImageNet. Vision-language models like LLaVA, leveraging the CLIP vision encoder, could exhibit benefits in classification performance for challenging categories over CLIP alone.
Our findings are poised to offer valuable guidance for enhancing the robustness and reliability of CLIP models.
\end{abstract}


\section{Introduction}
\IEEEPARstart{L}{everaging} contrastive training to cohesively align images and text within a singular embedding domain, the CLIP model excels in delivering versatile zero-shot generalizations. This inherent proficiency enables CLIP to handle diverse tasks without the need for task-specific fine-tuning~\cite{radford2021learning, jia2021scaling}.  
Remarkably, CLIP models exhibit outstanding zero-shot classification capabilities, even without explicit training on the target dataset. Moreover, they demonstrate commendable robustness against challenging natural distributional shifts~\mbox{\cite{recht2019imagenet, wang2019learning, hendrycks2021many, barbu2019objectnet, hendrycks2021natural}}.
Gaining a deeper understanding of such behaviors in CLIP models is crucial for steering the future image-text foundational models. Contemporary research has delved into multiple facets of CLIP models. This encompasses areas such as dataset formulation~\cite{nguyen2022quality}, reproducibility in scaling laws~\cite{cherti2022reproducible}, strategies for fine-tuning~\cite{wortsman2022robust}, adversarial classification robustness~\cite{zhao2024evaluating} and nuances of the training distribution~\cite{fang2022data, shi2023effective}.

Motivated by previous work, we conduct an in-depth analysis of CLIP models, expanding our perspective beyond overall classification robustness. Our analysis includes several key dimensions: (1) robustness to visual factors, where we assess whether CLIP models can maintain performance when encountering variations such as pose, size, color, lighting, and occlusions; (2) out-of-distribution (OOD) detection, evaluating the models’ ability to identify instances with labels not present in the training distribution; (3) predictive uncertainty, examining whether CLIP models provide calibrated predictions that accurately reflect uncertainty under different testing conditions; (4) zero-shot retrieval, assessing the models’ capability to associate novel textual queries with relevant visual content; (5) 3D awareness, evaluating how well CLIP models handle 3D corruptions and maintain multi-view consistency; and (6) interaction between the vision and language encoders, investigating how these components influence classification robustness.
Within each of these dimensions, we analyze the impact of several crucial factors on CLIP’s behavior, including variations in training distribution, model architectures, dataset sizes, contrastive loss, fine-tuning, test-time prompts, and dataset curation. This comprehensive analysis provides a thorough assessment of both the strengths and limitations of CLIP models across these critical areas.

To this end, we evaluate $84$ zero-shot CLIP models with varying visual encoder architectures, training sources, and dataset sizes, as well as $44$ ImageNet fine-tuned CLIP models. To establish a baseline, we compare these models against $127$ ImageNet models without language-image pre-training.
We examine $10$ visual factors variations present in the ImageNet validation set~\cite{idrissi2022imagenet}, including object pose, lighting, and background, to assess models' visual factors-level robustness.
As for OOD detection, we employ ImageNet as an in-distribution (ID) set following~\cite{ming2022delving} and test on $5$ types of OOD scenarios.
Then, to investigate the predictive uncertainty, we use a set of canonical ImageNet distributions, such as texture, style, and perturbation shifts.
We evaluate the effectiveness of data curation methods on the aforementioned datasets.
Furthermore, we measure the 3D awareness of CLIP by geometric, semantic correspondence estimation as in~\cite{el2024probing} and robustness against 3D-related corruptions, such as near focus and motion blur~\cite{kar20223d}.
Lastly, to explore the interplay between the visual and text encoders of CLIP, we compare CLIP models with LLaVA~\cite{liu2024visual} in terms of classification performance on the challenging diffusion model-generated ImageNet-D~\cite{zhang2024imagenet}.

This article extends our previous conference paper~\cite{tu2023robust}, with the following major additions:
\textbf{(1)} The experiment scale has been expanded by including 25 recent zero-shot CLIP models trained on different subsets of DATACOMP~\cite{gadre2023datacomp}, allowing us to broaden the findings to the medium-to-low accuracy regime of CLIP models.
\textbf{(2)} An in-depth analysis is provided to uncover the impact of fine-tuning objectives on the shape-bias of CLIP models~(Section~\ref{sec:robust2}).
\textbf{(3)} The zero-shot retrieval capability of CLIP models is explored, highlighting the significance of training distribution as a key factor affecting performance trends~(Section~\ref{sec:retrieval}).
\textbf{(4)} A comprehensive study of fine-tuning methods, including parameter-efficient, standard, and contrastive fine-tuning, is presented~(Section~\ref{sec:ft}).
\textbf{(5)} A new OOD benchmark, NINOC, is added in our evaluation, which is ID-free and aggregates OOD classes from multiple existing datasets~(Section~\ref{sec:ood}).
\textbf{(6)} The 3D-awareness of CLIP models is evaluated by testing their performance on 3D correspondence estimation and robustness against 3D corruptions
~(Section~\ref{sec:3d}).
\textbf{(7)} The interaction between visual and language encoders is investigated from a classification perspective~(Section~\ref{sec:llava}).
\textbf{(8)} We extend the evaluation of dataset curation techniques to robustness-related tasks, including out-of-distribution (OOD) detection, calibration, visual factor-level
robustness, and 3D corruption~(Section~\ref{sec:dc}).

We summarize our evaluation dimensions, covering the task, model, and data levels, along with key findings in Table~\ref{tab:eval_sum}.

\begin{table*}[t]
\centering
\setlength\tabcolsep{10pt}
\caption{\wj{\textbf{Summary of evaluation dimensions and key findings.} Our evaluation covers three levels of analysis: {task-level}, {model-level}, and {data-level}, each addressing distinct aspects of model behavior and robustness.}}
\label{tab:eval_sum}
\begin{tabular}{p{0.05cm}p{3.4cm}p{4.6cm}p{7.1cm}}
\toprule
& \textbf{Evaluation Dimension} & \textbf{What We Evaluated} & \textbf{Key Findings} \\
\midrule
\multirow{10}{*}{\rotatebox[origin=c]{90}{\textbf{I. Task-Level}}}
& \makecell[l]{Visual Factor Robustness \\ (Sec.~\ref{sec:robust})} & Performance under visual variations (\textit{e.g.}, shape, texture, and size) & CLIPs outperform ImageNet models on six of ten factors; Training distribution impacts visual factor robustness of CLIP.\\

& \makecell[l]{OOD Detection \\ (Sec.~\ref{sec:ood})} & Novelty detection ability (\emph{e.g.}, NINCO) & Zero-shot CLIP is competitive with other models; Training distribution impacts detection accuracy.
\\

& \makecell[l]{Calibration \\ (Sec.~\ref{sec:cal})} & Alignment between prediction confidence and correctness & Fine-tuning raises calibration error; LP-FT and WiSE-FT recover with temperature scaling, while FLYP remains overconfident. Both data distribution and quantity play key roles. \\

& \makecell[l]{Retrieval \\ (Sec.~\ref{sec:retrieval})} & Image–text matching accuracy &  CLIP’s Zero-shot retrieval aligns with classification accuracy; data distribution and augmentation shape retrieval quality.\\

& \makecell[l]{3D Awareness \\ (Sec.~\ref{sec:3d})} & Robustness to 3D corruptions and correspondence matching & CNN-based CLIPs are more stable than ViT-based CLIP.  \\
\midrule
\multirow{8}{*}{\rotatebox[origin=c]{90}{\textbf{II. Model-Level}}}
& \makecell[l]{Vision–Language Interaction \\ (Sec.~\ref{sec:llava})} & CLIP vs. LLaVA on hard category splits & LLaVA outperforms CLIP on ambiguous sets; stronger LLMs (Vicuna $>$ Mistral) amplify gains. \\

& \makecell[l]{Training Paradigm Impact \\ (Sec.~\ref{sec:paradigm})} & CLIP, BLIP, SigLIP, ViTamin & No paradigm dominates. ViTamin is more robust to 3D corruptions; SigLIP improves robustness but weak on calibration. \\

& \makecell[l]{Prompt Sensitivity \\ (Sec.~\ref{sec:prompt})} & Prompt set size and quality & LLM prompts improve accuracy but may not benefit OOD detection or calibration, and no method consistently dominates.  \\

& \makecell[l]{Fine-Tuning Impact \\ (Sec.~\ref{sec:ft})} & Various of fine-tuning techniques & LP-FT and WiSE-FT are well-balanced. FLYP boosts accuracy but hurts calibration. PromptSRC preserves both. \\
\midrule
& \makecell[l]{Dataset Curation Effect \\ \textbf{III. Data-level} (Sec.~\ref{sec:dc})} & Filtering and data diversity & Data curation boosts classification, OOD, retrieval, and 3D robustness for ViT-CLIP, but not calibration.\\

\bottomrule
\end{tabular}
\end{table*}

\section{Related Work}

\noindent\textbf{Robustness.} Machine learning models should generalize from training distribution to novel testing environments~\cite{djolonga2021robustness, koh2021wilds, kirsch2022note, huang2024machine, huang2023harnessing, huang2023winning}.
One line of work has developed a theoretical framework to investigate model robustness~\cite{6287402}. Ben-David et al.~\cite{6287402} were the first to propose a generalization bound based on the VC dimension, which quantifies the difference in classifier error between source and target distributions using a divergence measure. Mansour et al.\cite{mansour2009domain} later expanded this analysis to accommodate more general loss functions, offering improved generalization bounds through Rademacher complexity.
To investigate such capability of deep models to various forms of test distributions, a commonly used approach is to introduce artificial transformations onto images, such as style transfer~\cite{geirhos2018imagenet}, corruptions and perturbations~\mbox{\cite{hendrycks2019benchmarking, mintun2021interaction}}. 
Moreover, many real-world datasets are introduced to assess model robustness under different natural distributional shifts~\mbox{\cite{recht2019imagenet, wang2019learning, hendrycks2021many, barbu2019objectnet, hendrycks2021natural, baek2022agreement}}. 
For instance, Idrissi \textit{et al.},~\cite{idrissi2022imagenet} proposes ImageNet-X by relabelling the ImageNet validation set to provide detailed labels for naturally occurring factors such as pose, background, and lighting.~\cite{zhang2024imagenet} introduces 3DCC to study the robustness of networks to 3D corruptions.

\vspace{0.2cm}
\noindent\textbf{CLIP Analysis.} Existing studies have explored various aspects of CLIP models, including dataset formulation~\cite{nguyen2022quality}, reproducibility in scaling laws~\cite{cherti2022reproducible}, adversarial classification robustness~\cite{zhao2024evaluating}, fine-tuning strategies~\cite{wortsman2022robust}, nuances of the training distribution~\cite{fang2022data}, visual prompt~\cite{shtedritski2023does}, typographic attacks~\cite{cheng2024typographic} and techniques for dataset curation~\cite{gadre2023datacomp}. 

Our comprehensive evaluation of CLIP goes beyond overall classification robustness to include assessments of visual-factor robustness and 3D corruption robustness. We also explore additional perspectives that are crucial for real-world applications, such as out-of-distribution (OOD) detection, which aims to filter out inputs that are irrelevant to the task at hand. Furthermore, we examine prediction uncertainty to determine whether the model can classify images with calibrated prediction probabilities that align with the empirical frequency of correctness~\cite{guo2017calibration, nguyen2015posterior}. Additionally, we incorporate zero-shot retrieval tasks~\cite{cherti2022reproducible} and 3D geometry correspondence matching to investigate the potential of CLIP features.

\section{Experimental Setup}\label{sec4}

\subsection{Models of Interest} 

\noindent\textbf{Contrastive language-image pre-training models:} 
we use~\textbf{$84$ zero-shot CLIP models (CLIP)} and \textbf{$44$ ImageNet fine-tuned CLIP models (CLIP-FT)}. They have different visual encoders, including slightly modified ResNet~\cite{he2016deep}, ConvNeXt~\cite{liu2022convnet}, ViT~\cite{dosovitskiy2020image} and EVA~\cite{sun2023eva}. 
There are various training sources, including LAION~\cite{schuhmann2022laionb}, WIT~\cite{radford2021learning} and Conceptual Captions~\cite{sharma2018conceptual}, and multiple sizes of training datasets from {$3$} million to $2$ billion. 
Note that in this extended paper, we include $25$ recent zero-shot CLIP models. They are trained on subsets of CommonPool~\cite{gadre2023datacomp}, ranging from $14$ million, $140$ million to $1$ billion. CommonPool draws its data from the same source as LAION, which is Common Crawl.
These models allow us to validate and expand our findings in a medium-to-low accuracy regime. 
We also assess the performance of very recent CLIP models which are trained on filtered high-quality pre-training datasets using dataset curation techniques~\mbox{\cite{fang2023data, xu2023demystifying}}.
To compare the performance with LLaVA~\cite{liu2024visual}, we also include SigLIP~\cite{zhai2023sigmoid}.

For the CLIP-FT models, the vision encoder of CLIP is fine-tuned on ImageNet-1K.
{We consider different fine-tuning algorithms, including directly fine-tuned on ImageNet-1K~\cite{imagenet_cvpr09}, first fine-tuned on ImageNet-12K, a subset of ImageNet-22K before fine-tuning on ImageNet-1K, and also fine-tuned by parameter-efficient fine-tuning methods~\cite{zhou2022learning, zhang2022tip}}.
We use the default prompt template provided by~\cite{radford2021learning} for zero-shot CLIP models unless specified.

\vspace{0.2cm}
\noindent\textbf{Models compared:} 
we use $127$ ImageNet models with various architectures, including Convolutional Neural Networks (\emph{e.g.}, ResNet~\cite{he2016deep} and ConvNeXt~\cite{liu2022convnet}), Vision Transformers (\emph{e.g.}, ViT~\cite{dosovitskiy2020image} and Swin~\cite{liu2021swin}) and all-MLP architectures~\mbox{\cite{ding2021repmlp, tolstikhin2021mlp}} (\emph{e.g.}, MLP-Mixer~\cite{tolstikhin2021mlp}). 
Following~\cite{taori2020measuring}, we divide them into three categories: \textbf{(i) Standard Models.} This group consists of models supervised on the ImageNet training set. 
\textbf{(ii) Contrastive learning models.} This category contains $8$ models pre-trained by contrastive learning. There are $6$ training algorithms investigated, including InsDis~\cite{wu2018unsupervised}, MoCo~\cite{he2020momentum}, SimCLR~\cite{chen2020simple};
\textbf{(iii) Pre-trained on more data.} This group contains models pre-trained on a significantly larger dataset (\emph{e.g.}, ImageNet-21K) than the ImageNet training set. All the above models, including CLIP, are publicly available on TIMM~\cite{rw2019timm}, OpenCLIP~\cite{ilharco_gabriel_2021_5143773}.

\vspace{0.2cm}
\noindent\textbf{Modern vision language models:}
This paper considers LLaVA~\cite{liu2024visual}, which combines a frozen CLIP vision encoder and a large language model (\textit{e.g.}, Vicuna) for general-purpose visual and language understanding. In our study, we consider six LLaVA models: the visual encoders used are CLIP-L/14-336 and SigLIP, paired with three large language models: Mistral-instruct-V2~\cite{jiang2023mistral}, Llama-Chat~\cite{touvron2023llama}, and Vicuna-V2-7B~\cite{touvron2023llama}, resulting in a total of six LLaVA models. These models are available on HuggingFace, as provided by~\cite{karamcheti2024prismatic}.

\subsection{Test Sets and Metrics}

\noindent\textbf{I. Robustness.} 
We first pinpoint model failure patterns by testing on ImageNet-X~\cite{idrissi2022imagenet}, which is a relabeling of ImageNet validation by $16$ naturally occurring factors. This work mainly considers $10$ factors labelled with a sufficient number of test samples: \textit{Pose}, \textit{Background}, \textit{Pattern}, \textit{Color}, \textit{Smaller}, \textit{Shape}, \textit{Partial View}, \textit{Subcategory}, \textit{Texture} and \textit{Larger}.
The metric is accuracy, and high is better. In addition, we include cue-conflict stimuli and Stylized-ImageNet~\cite{geirhos2018imagenet} to measure the model bias towards the shape or texture.

\vspace{0.2cm}
\noindent\textbf{II. OOD detection.} 
We use a large-scale OOD detection benchmark which is built up on ImageNet: in-distribution (ID) ImageNet \emph{v.s.} $\{$iNaturalist~\cite{van2018inaturalist}, SUN~\cite{xiao2010sun}, PLACES~\cite{zhou2017places}, TEXTURE~\cite{cimpoi2014describing}, and ImageNet-O~\cite{hendrycks2021natural} (OOD).  Metrics are the area under the receiver operating characteristic curve (AUROC) and the higher is better; false positive rate (FPR@95) when the true positive rate is at 95\% and a lower score is better. 
To evaluate OOD detection across diverse conditions, we employ the NINCO dataset~\cite{bitterwolf2023or}, which is ID-contamination-free and comprises OOD classes from various existing OOD datasets. We report mean AUROC and FPR@95.

\vspace{0.2cm}
\noindent\textbf{III. Calibration.} 
We study ID and OOD datasets, where ImageNet validation is ID dataset and OOD datasets are: ImageNet-V2~\cite{recht2019imagenet}, ImageNet-Rendition~\cite{hendrycks2021many}, ImageNet-Adversarial~\cite{hendrycks2021natural}, ImageNet-Sketch~\cite{wang2019learning}, ObjectNet~\cite{barbu2019objectnet} and ImageNet-Vid-Robust~\cite{shankar2021image}. Metrics are estimated calibration error (ECE)~\cite{naeini2015obtaining} and negative log-likelihood (NLL). A lower ECE or NLL indicates better calibration.

\vspace{0.2cm}
\noindent\textbf{IV. Retrieval.} We evaluate zero-shot retrieval performance on Flick30K~\cite{young-etal-2014-image} and MSCOCO~\cite{chen2015microsoft} following the evaluation setup and splits from~\cite{karpathy2015deep}.
As in~\cite{radford2021learning}, we compute the cosine similarity between image and text embeddings as the image-text scores. When evaluating image retrieval, we rank the top-$K$ images for each text caption, and vice versa for text retrieval. Recall$@K$ is the metric with $K = 5$. 

\vspace{0.2cm}
\noindent\textbf{V. 3D Awareness.} Two tasks are explored for this property: correspondence estimation and robustness against 3D corruptions. We use ScanNet~\cite{dai2017scannet}, NAVI~\cite{jampani2023navi} and SPair-71K~\cite{min2019spair} as the evaluation datasets for correspondence estimation. The metric is recall. For robustness against 3D corruptions, we use 3DCC~\cite{zhang2024imagenet}, which applies 3D-related corruptions against ImageNet-validation with $5$ severity levels. The performance is measured by accuracy. 

\vspace{0.2cm}
\noindent\textbf{VI. Comparison to LLaVA.} We compare the performance of CLIP and LLaVA on ImageNet-D\cite{zhang2024imagenet}, which consists of three splits, \textit{Background}, \textit{Texture} and \textit{Material}. CLIP is evaluated using standard zero-shot image classification protocol, while LLaVA is assessed by standard visual question answering protocol. They are both required to classify images from four classes and use accuracy as the measurement.

\begin{figure*}[t]
    \centering
    \includegraphics[width=\linewidth]{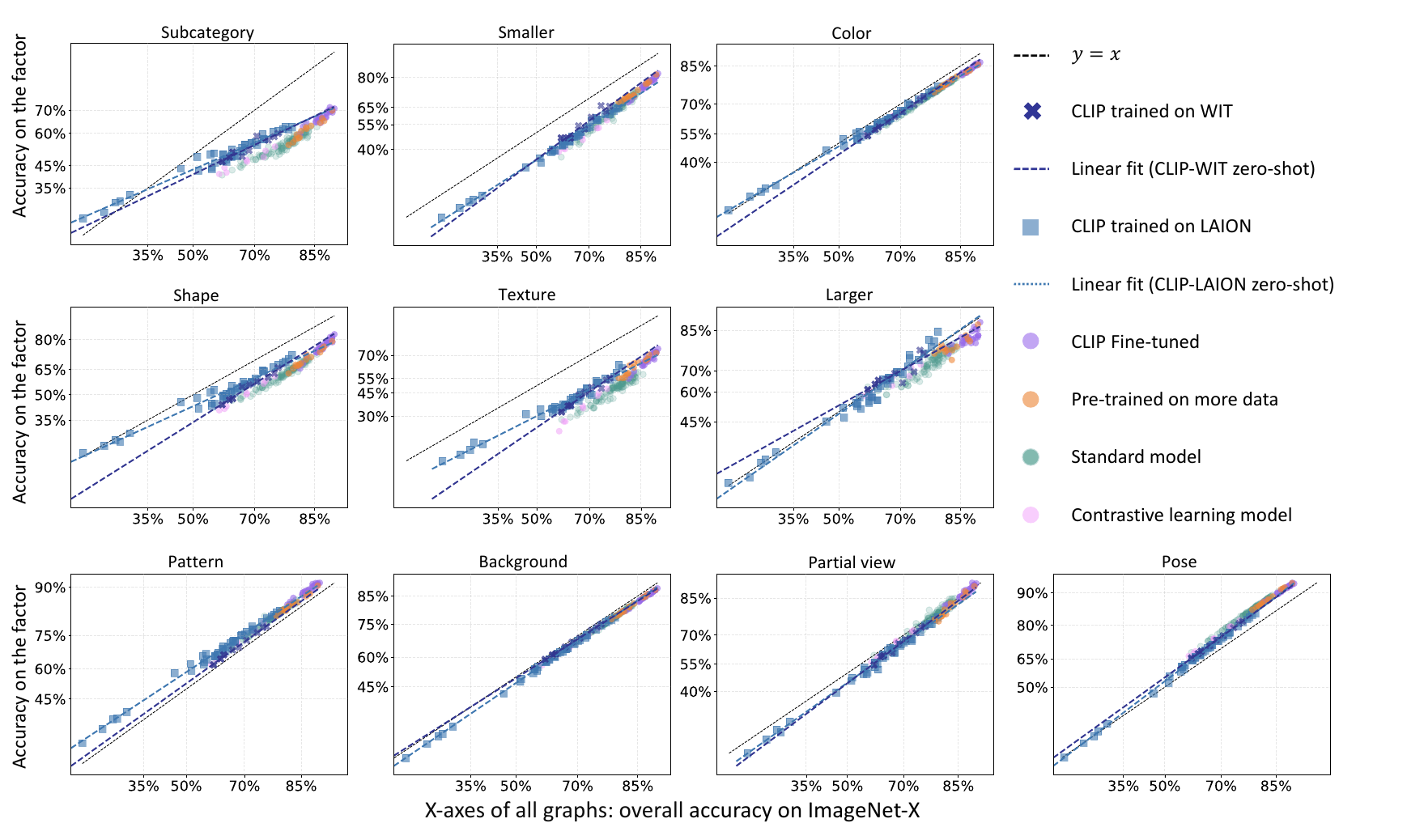}
    \caption{
    \textbf{The models' performance on the subset of ImageNet-X annotated with a given visual factor (y-axis) to their overall accuracy on the whole ImageNet-X (x-axis).} Each point represents a model. The x-axis and y-axis are probit transformed following~\cite{taori2020measuring}. The black dashed line represents the ideal robust models whose performance on each visual factor is the same as the overall performance. The blue straight lines are fit with robust linear regression~\cite{huber2011robust}.
    We include models supervised on ImageNet-1K, pre-trained on more data, contrastive learning models, CLIP models trained on two data distributions, and their fine-tuned counterparts. 
    }
    \label{fig:factor}
\end{figure*}

\subsection{Analytical Methodology}

\textbf{Key Factors.} to understand the underlying factors that influence the performance of CLIP models, we delve into six primary aspects: 1) training distribution, evaluating the effect of data source; 2) model architecture, looking into the potential effects of different structural choices on model performance; 3) dataset quantity, probing the interplay between the amount of data available for training and the model's efficiency; 4) contrastive loss, understanding its specific role in training dynamics 5) fine-tuning, 6) test-time prompt, assessing the impact of prompts during the evaluation on model outputs.

We follow the analytical methodology of seminal work~\cite{taori2020measuring}, along with subsequent studies such as~\mbox{\cite{nguyen2022quality, fang2022data, miller2021accuracy}}, to study the influential factor. 
Within the performance trends observed across all models, any factor causing a deviation from these trends is influential. Notably, in our research, we mainly emphasize and discuss such influential factors within each facet of our investigation.
In Table~\ref{tab:eval_sum}, we organize our evaluation into task-level, model-level, and data-level dimensions, highlighting the main insights observed in each.

\section{Visual Factor-Level Robustness} \label{sec:robust}
Our research builds upon previous findings on the robustness of CLIP models and focuses on the potential failure types of the model. Instead of solely measuring overall accuracy across distributions, this section investigates the behavior of CLIP models when faced with varying visual factors such as \textit{Pose}, \textit{Background}, and \textit{Object Scale}. 

\subsection{CLIP Models Generally Exhibit Better Factor-Level Robustness Than Other Models}
\label{sec:robust1}
\vspace{0.2cm}
\noindent\textbf{Factor-level effective robustness.} 
In our study, we introduce the concept of visual factor-level effective robustness based on effective robustness~\cite{taori2020measuring}. It measures a model's ability to achieve higher accuracy on the subset annotated by a specific visual factor compared to what is expected based on its overall accuracy on ImageNet-X. Fig.~\ref{fig:factor} displays the accuracy on the subset annotated by a specific visual factor relative to the overall accuracy on ImageNet-X.

\vspace{0.2cm}
\noindent\textbf{(1) CLIP models are generally more robust than other ImageNet models on six out of ten visual factors.} 
Fig.~\ref{fig:factor} highlights several insights into the factor-level robustness of CLIP models. First, we find that CLIP models are more robust than other models on six out of ten visual factors, including \textit{Subcategory}, \textit{Smaller}, \textit{Color}, \textit{Shape}, \textit{Texture}, and \textit{Larger}. 
{Specifically, CLIP models exhibit higher factor-level effective robustness than other models on each of these factors. }
Second, we observe that CLIP models are less robust than other models on \textit{Pose} and \textit{Partial View}. Third, CLIP models show a similar trend to other models on the \textit{Background} factor. 

\vspace{0.2cm}
\noindent\textbf{(2) Training distributions lead to different trends in CLIP models.} 
The choice of training distribution impacts the factor-level robustness of CLIP models. Specifically, we find that training on different datasets (\emph{i.e.}, LAION and WIT) forms distinct trends on each visual factor for CLIP, and there is no single training source that always leads to higher factor-level robustness than another.
For instance, we observe that CLIP models trained on LAION demonstrate higher robustness on \textit{Shape} factor than those trained on WIT, while this reverses for \textit{Background} and \textit{Pose} factors. The results show a mixed observation on \textit{Large} factor.
Furthermore, we further point out that CLIP models trained on different subsets of LAION (LAINON-80M, LAION-400M, and LAION-2B) follow the same trend. 
The above observations highlight the importance of the choice of training source in determining not only the overall accuracy but also the factor-level behaviors of CLIP models. This suggests that factor-level robustness should be considered when choosing the training source.

\vspace{0.2cm}
\noindent\textbf{(3) CLIP fine-tuned models perform slightly better than models pre-trained with more data}. 
We compare CLIP fine-tuned models (CLIP-FT) with other models pre-trained on more data and find that CLIP-FT shows improvement in overall accuracy and robustness on visual factors of \textit{Subcategory}, \textit{Shape}, and \textit{Pattern}. However, no additional robustness gain is observed on other factors. 
Moreover, CLIP-FT models outperform zero-shot CLIP on variations such as \textit{Pattern} and \textit{Partial View} but perform lower on factors like \textit{Texture} and \textit{Larger}. We speculate that standard fine-tuning introduces spurious correlations~\cite{xiao2023masked}. This may lead to a bias for CLIP towards specific visual properties, thereby compromising factor-level robustness on some factors. It would be intriguing to explore fine-tuning techniques to maintain or improve the visual factor-level robustness of CLIP.

\noindent\wj{\noindent\textbf{(4) Discussion on consistent trends across visual factors.}
All models exhibit consistent trends across visual factors, despite differences in architecture and training data. Specifically, all models lie below the line $y = x$ under \textit{Smaller}, \textit{Shape}, and \textit{Texture} conditions, which involve changes to object geometry, scale, and surface patterns. While such variations do occur in natural datasets, they are neither explicitly annotated nor emphasized, and thus may be underrepresented in the models’ learned feature space. As a result, models tend to rely on statistically dominant but fragile cues—such as canonical shapes, common textures, or typical object sizes—rather than learning representations that are robust to these factors. This behavior is consistent with the concept of shortcut learning~\cite{geirhos2020shortcut}, where models exploit superficial but predictive patterns that fail under distribution shift. In contrast, performance on Background and Partial View remains stable, likely due to the abundance of such variations in pretraining data, which encourages models to downweight context and develop object-centric representations. The consistency across models suggests these are not architecture-specific artifacts but shared limitations shaped by training data and objectives.}

\noindent\wj{\noindent\textbf{(5) Pre-training analysis of factor-level coverage for training data selection}.
Given a candidate training set, we aim to estimate its coverage across key visual factors to anticipate potential robustness gaps before model training. 
ImageNet-X provides annotations across $16$ visual factors, which supports pre-training analysis. Specifically, we compute a prototypical feature for each factor by averaging features from its annotated images using a fixed pretrained backbone (\textit{e.g.}, ResNet-50). For each image in the candidate dataset, we extract its feature and assign it to the nearest factor prototype. By counting how many training data are associated with each factor, we estimate the dataset's factor-level coverage. This lightweight and scalable analysis enables factor-aware data selection before training models.}

\begin{table}
    \centering
    \setlength\tabcolsep{15pt}
    \begin{tabular}{c cc}
    \toprule    
    Backbone & FT methods &  Shape bias \\ \midrule
    \multicolumn{1}{c}{\multirow{5}{*}{\makecell{ViT-B/32}}} & {Zero shot}  & {$0.575$} \\ 
     & {Fine-tune on 1K}  & {$0.401$} \\
    & {Contrastive FT}  & {$0.561$} \\
    & {CoOp}  & {$0.549$} \\
    & {Tip-Adapter}  & {${0.579}$} \\
    
    \midrule    
    \multicolumn{1}{c}{\multirow{5}{*}{\makecell{ViT-B/16}}} & {Zero shot}  & {$0.473$} \\ 
     & {Fine-tune on 1K}  & {$0.345$} \\
    & {Contrastive FT}  & {$0.448$} \\
    & {CoOp}  & {$0.472$} \\
    & {Tip-Adapter}  & {${0.487}$} \\
    \bottomrule 
    \end{tabular}
    \vspace{0.3cm}
    \caption{\textbf{Shape bias of various fine-tuned CLIP models.} We include CLIP models fine-tuned using different methods: cross-entropy, contrastive loss~\cite{goyal2023finetune}, and parameter-efficient techniques such as CoOp~\cite{zhou2022learning} and Tip-Adapter~\cite{zhang2022tip}.}    %
    \label{tab:shape-bias}
\end{table}

\begin{figure*}[t!]
\begin{minipage}{0.4\linewidth}
    \label{fig:ablation}
    \includegraphics[width=\linewidth]{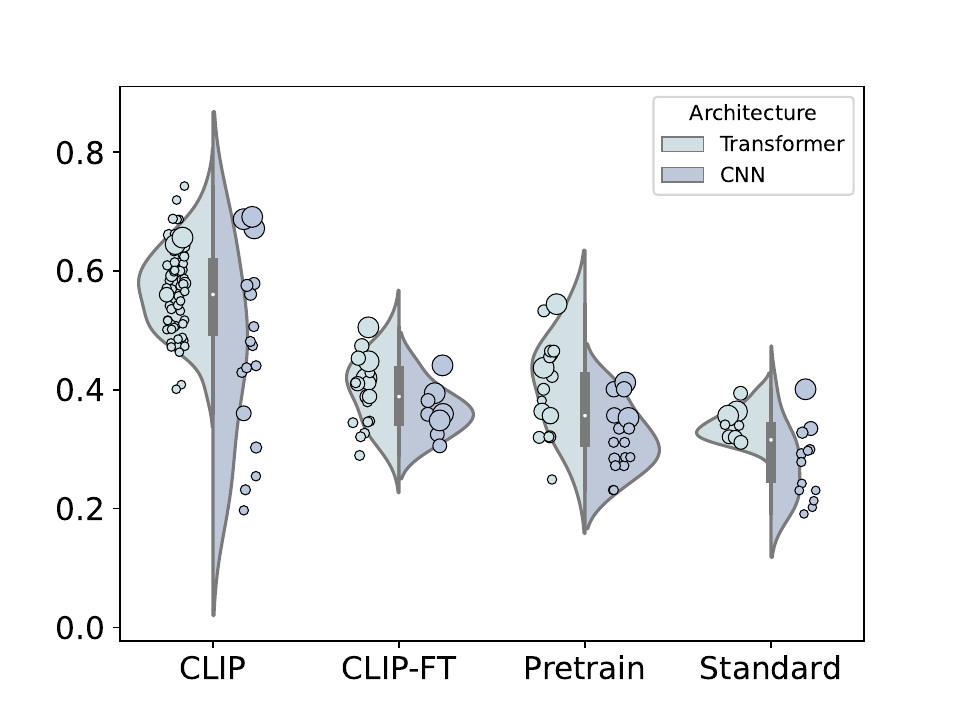}
    \figcaption{\textbf{Shape bias analysis} of CLIP, CLIP fine-tuned (CLIP-FT), models pre-trained on more data (Pretrain), and standard models. Large points mean larger models within the group. We observe that CLIP models are more shape-biased.}
     \label{fig:cue}
\end{minipage}
\:
\begin{minipage}{0.58\linewidth}
\centering
\vspace{0.8cm}
\begin{adjustbox}{width=\linewidth}
    \begin{tabular}{c cccc}
    \toprule    
    Source & Backbone &  Shape bias & IN-Val & SIN \\ \midrule
    \multicolumn{1}{c}{\multirow{5}{*}{\makecell{LAION}}} & {ViT/H-14 (336/224)}  & {$0.42$ /$\mathbf{0.51}$} & {$\mathbf{0.89}$ /$0.88$} & {$0.28$ /$\mathbf{0.32}$}\\ 
     & {ViT/L-14 (336/224)}  & {$0.41$ /$\mathbf{0.47}$} & {$\mathbf{0.88}$ /$\mathbf{0.88}$} & {$0.27$ /$\mathbf{0.31}$}\\
    & {ViT/B-16 (384/224)}  & {$0.35$ /$\mathbf{0.43}$} & {$\mathbf{0.87}$ /$0.86$} & {$0.23$ /$\mathbf{0.25}$}\\
    & {ViT/B-32 (384/224)}  & {$0.33$ /$\mathbf{0.45}$} & {$\mathbf{0.85}$ /$0.83$} & {$0.21$ /$\mathbf{0.22}$}\\
    & {ConvNeXt-B (384/224)}  & {$0.31$ /$\mathbf{0.38}$} & {$\mathbf{0.87}$ /$0.86$} & {$0.17$ /$\mathbf{0.21}$}\\
    
    \midrule    
    \multicolumn{1}{c}{\multirow{2}{*}{\makecell{WIT}}}
     & {ViT/L-14 (336/224)}  & {$0.39$ /$\mathbf{0.45}$} & {$\mathbf{0.88}$ /$\mathbf{0.88}$} & {$0.24$ /$\mathbf{0.30}$}\\
    & {ViT/B-16 (384/224)}  & {$0.35$ /$\mathbf{0.41}$} & {$\mathbf{0.87}$ /$0.86$} & {$0.22$ /$\mathbf{0.23}$}\\
    
    \bottomrule 
    \end{tabular}
	\end{adjustbox}
	  \vspace{0.8cm}
	\tabcaption{\textbf{The influence of input resolution on shape bias when fine-tuning CLIP. We also report accuracy on ImageNet-Val(idation) and Stylized ImageNet (SIN).} The higher value in a model pair is in bold. With the same backbone architecture, the CLIP model fine-tuned with a larger input resolution is more accurate on IN-Val but less shape-biased and less accurate on SIN.}
\label{tab:cue}
\end{minipage}
\end{figure*}

\subsection{Texture Bias \textit{v.s.} Shape Bias}
\label{sec:robust2}

\noindent\textbf{CLIP exhibits a shape bias.} 
We conducted experiments using the cue-conflict stimuli dataset~\cite{geirhos2018imagenet} to assess the presence of shape bias in the model’s predictions. Shape bias, in this context, refers to the proportion of correct predictions that are based on the object’s shape rather than texture or other features. Fig.~\ref{fig:cue} visualizes the shape bias exhibited by different models, grouped by training methods (zero-shot, CLIP fine-tuning, additional data pre-training, and standard training) and architecture (transformer versus CNN).
Our results show that, among the four training methods, CLIP models exhibit a stronger shape bias compared to the other groups. While previous research has indicated that transformers show a greater shape bias than CNNs~\cite{naseer2021intriguing, zhang2022delving}, we found that CLIP models with CNN-based vision encoders also exhibit a significant shape bias. This suggests that CLIP can align more closely with human visual perception, which is widely acknowledged to be shape-driven~\cite{geirhos2018imagenet, hermann2020origins,li2024emergence}.
In the following, we provide a more detailed analysis of the shape bias observed in CLIP models and explore the implications of these findings.

\vspace{0.2cm}
\noindent\textbf{(1) Model size does not solely explain the shape bias of CLIP.}
We further observe that larger CLIP models do not necessarily have higher shape bias than smaller-size ones. For example, two models both trained on LAION-80M, CLIP-ViT/L-14 have $0.54$ shape bias, which is $0.09$ lower than CLIP-ViT/B-32. This implies that the shape bias of CLIP models cannot be attributed solely to model size. Based on the above, we speculate that the shape bias of CLIP may be attributed to its objective, which involves training the model to associate text and image pairs.

\vspace{0.2cm}
\noindent\textbf{(2) Larger input image resolution during fine-tuning of CLIP results in a stronger bias towards texture.} 
In Table~\ref{tab:cue}, we observe that an input resolution during fine-tuning impacts shape bias: increasing input resolution during fine-tuning leads to better accuracy on ImageNet validation but also results in more texture-biased models with lower accuracy on Stylized-ImageNet. Across seven pairs of experiments and two training sources, we observe this pattern consistently. 
Given that input resolution is a crucial model dimension~\cite{tan2019efficientnet,bello2021revisiting,tan2021efficientnetv2}, it would be insightful to study its effects on shape bias beyond classification accuracy when devising scaling strategies.

\vspace{0.2cm}
\noindent\textbf{(3) CLIP models tend to texture bias after fine-tuning.}
Our study reveals that shape bias in CLIP weakens after fine-tuning on ImageNet. Moreover, the fine-tuned CLIP models exhibit a shape bias comparable to models that are pre-trained on larger datasets. This finding is consistent when using a transformer and CNN as the visual encoder. Moreover, these results illustrate that fine-tuning discards the shape-biased property of zero-shot CLIP, which may affect their overall effective robustness~\cite{geirhos2019inducing, geirhos2018imagenet}.

{\vspace{0.2cm}
\noindent\textbf{(4) Fine-tuning with contrastive loss maintains shape bias.}
By default, the CLIP-FT models are trained with standard supervised cross-entropy loss. 
To decouple the effect of fine-tuning methods and data source, we use zero-shot CLIP with ViT-B/32 and ViT-B/16, and fine-tune them on ImageNet training set by standard cross-entropy,  contrastive loss~\cite{goyal2023finetune}, and parameter-efficient fine-tuning methods (CoOp~\cite{zhou2022learning} and Tip-Adapter~\cite{zhang2022tip}). The shape bias extents are shown in Table~\ref{tab:shape-bias}: contrastive fine-tuning on ImageNet maintains the shape bias of CLIP models. This indicates that ImageNet training data might not be the primary cause of the shape-bias decrease. \wj{We believe that the alignment mechanism between visual and textual representations may play a fundamental role in shaping this bias. This is supported by our observation that fine-tuning strategies which preserve image–text embedding association tend to retain or strengthen shape bias.}

\section{Out-of-Distribution Detection}
\label{sec:ood}

\begin{figure*}[t]
    \centering
    \includegraphics[width=\linewidth]{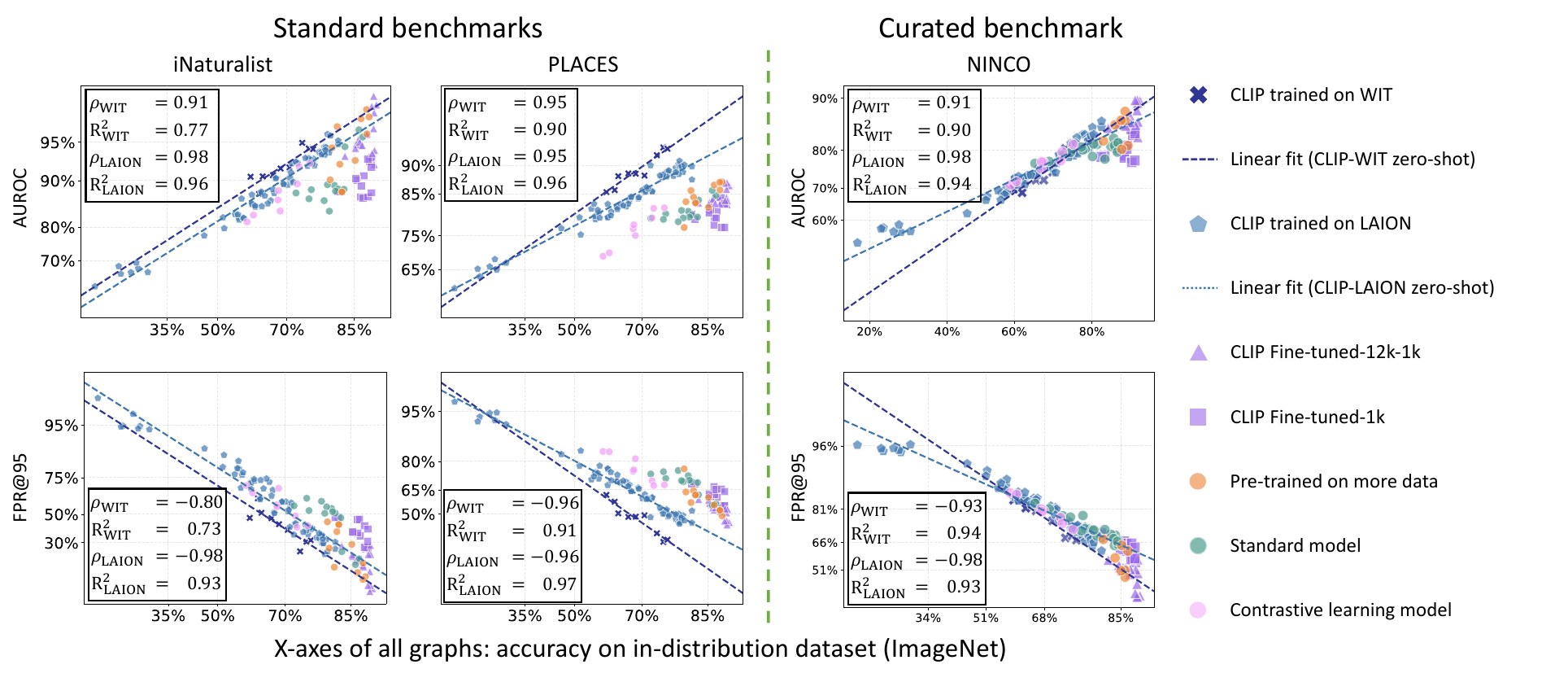}
    \caption{
    \textbf{OOD sample identification capability of models \emph{vs.} ID dataset classification accuracy.} The OOD detection ability is measured by AUROC ($\uparrow$) and FPR@95 ($\downarrow$). Each point represents a model. We plot the results on iNaturalist, PLACES and NINCO. The blue straight lines are fit with robust linear regression~\cite{huber2011robust}. \wj{We report spearman's rank correlation and $R^2$ to quantify the correlation strength between ID accuracy and OOD detection performance for zero-shot CLIP trained on WIT and LAION}. The x-axis and y-axis are probit transformed following~\cite{taori2020measuring}.
    We observe that training distribution has a greater impact than training dataset quantity on the OOD detection performance of CLIP. Moreover, after additionally fine-tuning on ImageNet-12K, CLIP models are generally better at detecting OOD samples than those only fine-tuned on ImageNet-1K.
    }
    \label{fig:ood}
\end{figure*}

Zero-shot CLIP allows for a flexible definition of in-distribution (ID) classes without re-training the model. Namely, they can conduct zero-shot OOD detection~\cite{ming2022delving}. The current findings suggest that zero-shot CLIP models are competitive with other state-of-the-art models~\cite{ming2022delving, fort2021exploring}. 
Based on this finding, we conduct an extensive analysis to determine whether the purported benefits persist across various training sources, subsets, and network architectures.
In the experiments, for zero-shot CLIP models, we utilize maximum concept matching~\cite{ming2022delving} to detect OOD data. For models that are trained or fine-tuned on ImageNet-1K, we employ maximum softmax score~\cite{hendrycks2016baseline} for OOD detection.

\vspace{0.2cm}
\noindent\textbf{(1) For CLIP models from the same source, their ID accuracy correlates with OOD detection performance.} 
Our study includes CLIP models trained on two sources (WIT and LAION). Given the same training source, our study, conducted across five challenging OOD scenarios, reveals a strong correlation between the ID accuracy of zero-shot CLIP models and their OOD detection performance (measured by AUROC and FPR@95). 
This suggests that the zero-shot classification accuracy of CLIP on ID data can serve as a reliable indicator of their OOD detection performance. 
In contrast, such a trend is not as strong for both standard models and more data-pre-trained models. Furthermore, CLIP-FT models fine-tuned on ImageNet-1K do not exhibit such a clear correlation.

\vspace{0.2cm}
\noindent\textbf{(2) Training source impacts the trend of CLIP.}
Upon closer examination of the training distribution, we have observed that the correlation trend between ID accuracy and OOD detection performance is largely dependent on the training source. As illustrated in Fig.~\ref{fig:ood}, our research shows two distinct trends between CLIP models trained on WIT and those trained on LAION.
Moreover, with the same ID accuracy, CLIP models trained on WIT exhibit superior OOD detection performance compared to their counterparts trained on LAION on three OOD scenarios. This further indicates the importance of training sources for CLIP. 

\vspace{0.2cm}
\noindent\textbf{(3) Fine-tuning procedure significantly influences the OOD detection ability of CLIP.} 
While fine-tuning generally improves CLIP's classification performance, this enhancement does not necessarily translate to better OOD detection accuracy. Some fine-tuned CLIP (CLIP-FT) models perform worse in OOD detection compared to their zero-shot counterparts.
Our analysis distinguishes between two groups of CLIP-FT models based on their fine-tuning procedures: one group is fine-tuned solely on ImageNet-1K, while the other undergoes additional fine-tuning on \mbox{ImageNet-12K}. We observe that this additional fine-tuning step has a notable impact on OOD detection performance. As shown in Fig.~\ref{fig:ood}, despite not yielding significant gains in classification accuracy, CLIP-FT models fine-tuned on \mbox{ImageNet-12K} consistently achieve better OOD detection across all tested scenarios.
These findings suggest that the fine-tuning dataset plays a critical role in enhancing OOD detection. Future work should further explore alternative fine-tuning strategies that prioritize OOD detection performance. Additionally, investigating the effects of fine-tuning on datasets beyond ImageNet-1K/21K presents an intriguing direction for improving the robustness of CLIP models.

\vspace{0.2cm}
\noindent\textbf{(4) Evaluation on NINCO~\cite{bitterwolf2023or}.} 
To explore the OOD detection across diverse and challenging conditions, we use a new benchmark NINCO for study. It consists of filtered samples from various existing OOD benchmarks.
Fig.~\ref{fig:ood} illustrates the OOD detection performance on NINCO versus ID classification accuracy on the ImageNet validation set. The observations are consistent with those on five standard benchmarks: 1) for CLIP models from the same source, their ID accuracy correlates with OOD detection; 2) training source influences trends of CLIP; 3) additional fine-tuning on ImageNet-12K helps OOD detection ability of CLIP.
\wj{ImageNet-21K offers broader semantic coverage than ImageNet-1K, which may help bridge the gap between pretraining data (\textit{e.g.}, LAION) and downstream tasks. As an intermediate fine-tuning stage, it could help preserve model generalization, which may explain the improved OOD robustness observed compared to direct fine-tuning on ImageNet-1K.}

\section{Confidence Calibration}
\label{sec:cal}
To better understand the well-calibrated phenomenon of zero-shot CLIP models reported by~\cite{minderer2021revisiting}, this section systematically analyzes the calibration behavior of CLIP models under various training conditions. Specifically, we examine the calibration performance of CLIP models trained on different training distributions, varied training set sizes, and different architectures. Furthermore, we also investigate the calibration performance of CLIP models after fine-tuning.

\begin{figure*}[t]
    \centering
    \includegraphics[width=0.95\linewidth]{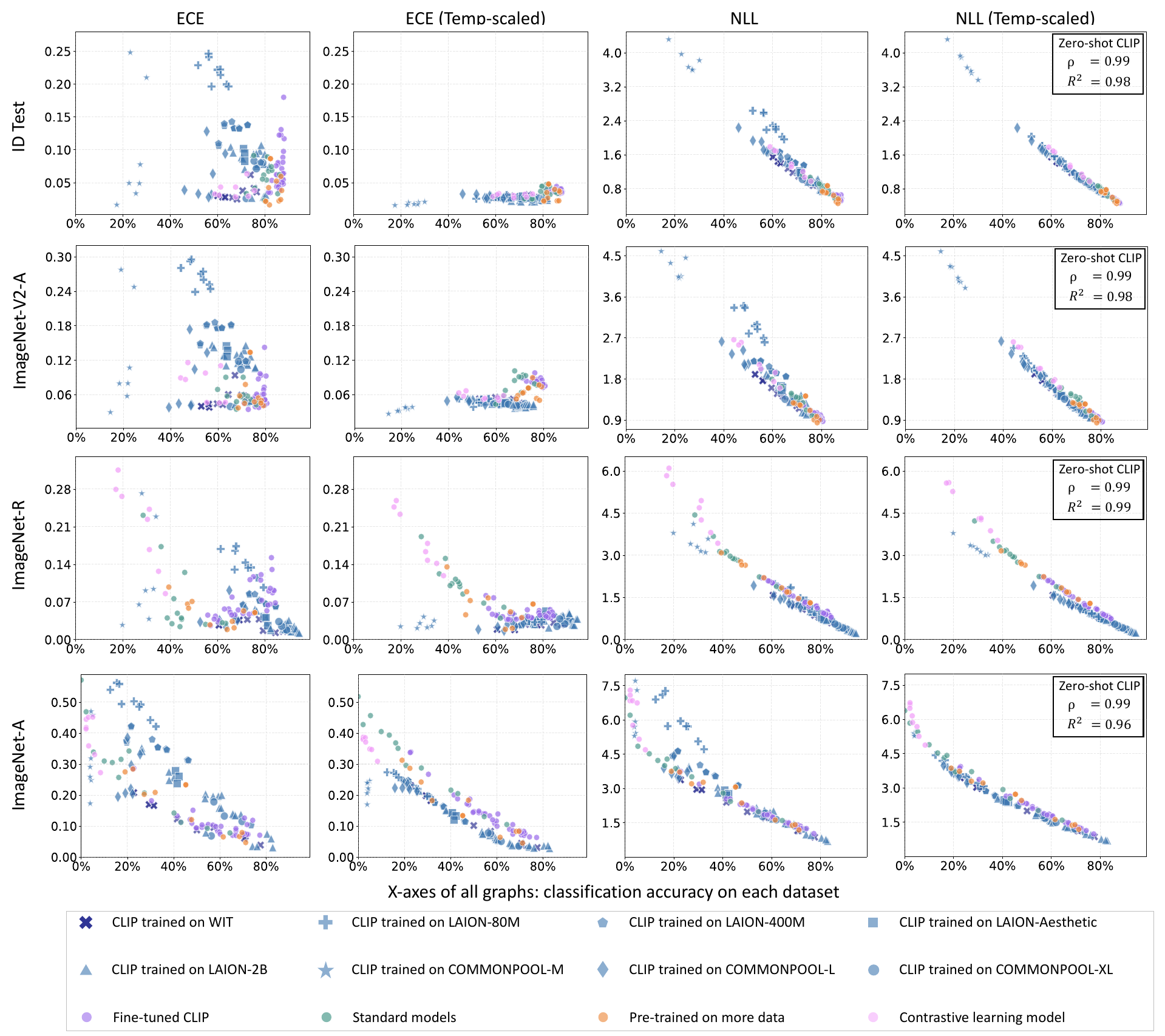}
    \caption{
    \textbf{Model calibration performance with respect to classification accuracy}. We report results on in-distribution test set, ImageNet-V2-A, ImageNet-R, and ImageNet-A. Two metrics are considered: ECE ($\downarrow$) and NLL ($\downarrow$), we also include calibration performance after calibration with temperature scaling. Each point represents a model. We use colors to represent model groups. For zero-shot CLIP, we additionally use shapes to indicate training distribution and quantity.
    CLIP models can have higher ECE than standard models. Also, the training distribution and quantity are the key factors influencing the calibration performance of CLIP models. Moreover, temperature scaling reveals a consistent trend in CLIP models. 
    After using temperature scaling for both CLIP and other models, CLIP models follow a distinct trend from others and show better calibration performance
    }
    \label{fig:cal}
\end{figure*}

\subsection{Zero-Shot CLIP Models Are Not Consistently More Calibrated Than Other Models}
\noindent\textbf{(1) Training data distribution and quantity significantly affect CLIP’s calibration.}
Fig.~\ref{fig:cal} illustrates the calibration of CLIP models concerning classification accuracy under distribution shifts. We find that models trained on different distributions or dataset sizes do not always group consistently. For example, CLIP models trained on WIT and LAION tend to form distinct clusters. Additionally, within subsets of the LAION dataset, models with similar classification accuracy can display varying levels of calibration.
While CLIP models are often praised for superior calibration compared to other models~\cite{minderer2021revisiting}, our analysis shows this is not always the case. Notably, CLIP models trained on the LAION-80M dataset exhibit significantly lower calibration performance compared to standard models.
The superior calibration reported by~\cite{minderer2021revisiting} is primarily based on CLIP models trained on WIT. However, when we expand the analysis to models trained on the broader LAION dataset and its subsets, we observe more variability.

\vspace{0.2cm}
\noindent\textbf{(2) CLIP fine-tuned models show a trade-off between calibration and classification.}
As shown in Fig.~\ref{fig:cal}, fine-tuning CLIP models consistently results in higher classification accuracy but increased calibration error across all test sets. 
Furthermore, we did not observe that further fine-tuning CLIP on ImageNet-12K benefits calibration performance, which contrasts with its positive impact on OOD detection.
Interestingly, other model groups, including those pre-trained on larger datasets, do not show an obvious trade-off between calibration and classification. Additionally, we observe that few fine-tuned CLIP models achieve better calibration than their zero-shot counterparts, even before applying calibration techniques.

\subsection{Temperature Scaling Highlights Well-Calibrated Properties of Zero-Shot CLIP Models}

Post-hoc calibration methods, such as temperature scaling~\cite{guo2017calibration}, are often employed to correct overconfidence or underconfidence in model predictions. Following the protocol in~\cite{gupta2021calibration}, we split the ImageNet validation set into two halves: one for temperature scaling (ID calibration) and the other for testing. We report results on both in-distribution~(ID) and out-of-distribution (OOD) test sets.

\vspace{0.2cm}
\noindent\textbf{(1) Classification accuracy of CLIP models correlates with calibration performance after temperature scaling.}
In Fig.~\ref{fig:cal}, we examine the effects of temperature scaling on both CLIP and non-CLIP models, grouped based on the amount and source of their training data. After applying temperature scaling and evaluating with the negative log-likelihood (NLL) metric, we observe that models with higher classification accuracy generally show better calibration. Importantly, when temperature scaling is applied to both CLIP and other models, zero-shot CLIP models consistently outperform other models, including fine-tuned versions, in calibration.

This pattern persists across various testing conditions, including ID and OOD sets, with zero-shot CLIP models demonstrating superior calibration compared to other models. This trend holds across both NLL and ECE metrics.

\vspace{0.2cm}
\noindent\textbf{(2) ID calibration of CLIP models transfers to OOD test sets.}
While prior studies~\cite{ovadia2019can} report in-distribution (ID) calibration often fails to generalize under distribution shifts, our findings reveal a promising result for CLIP models. After calibrating CLIP models on the ID set, they exhibit improved calibration on OOD test sets. For example, on ImageNet-A, CLIP models exhibit lower calibration error after temperature scaling, a trend not seen in other models. This suggests that CLIP models are relatively easier to calibrate across diverse distributions, indicating their potential for robust and reliable applications in real-world settings.

\begin{figure}
    \centering
    \includegraphics[width=\linewidth]{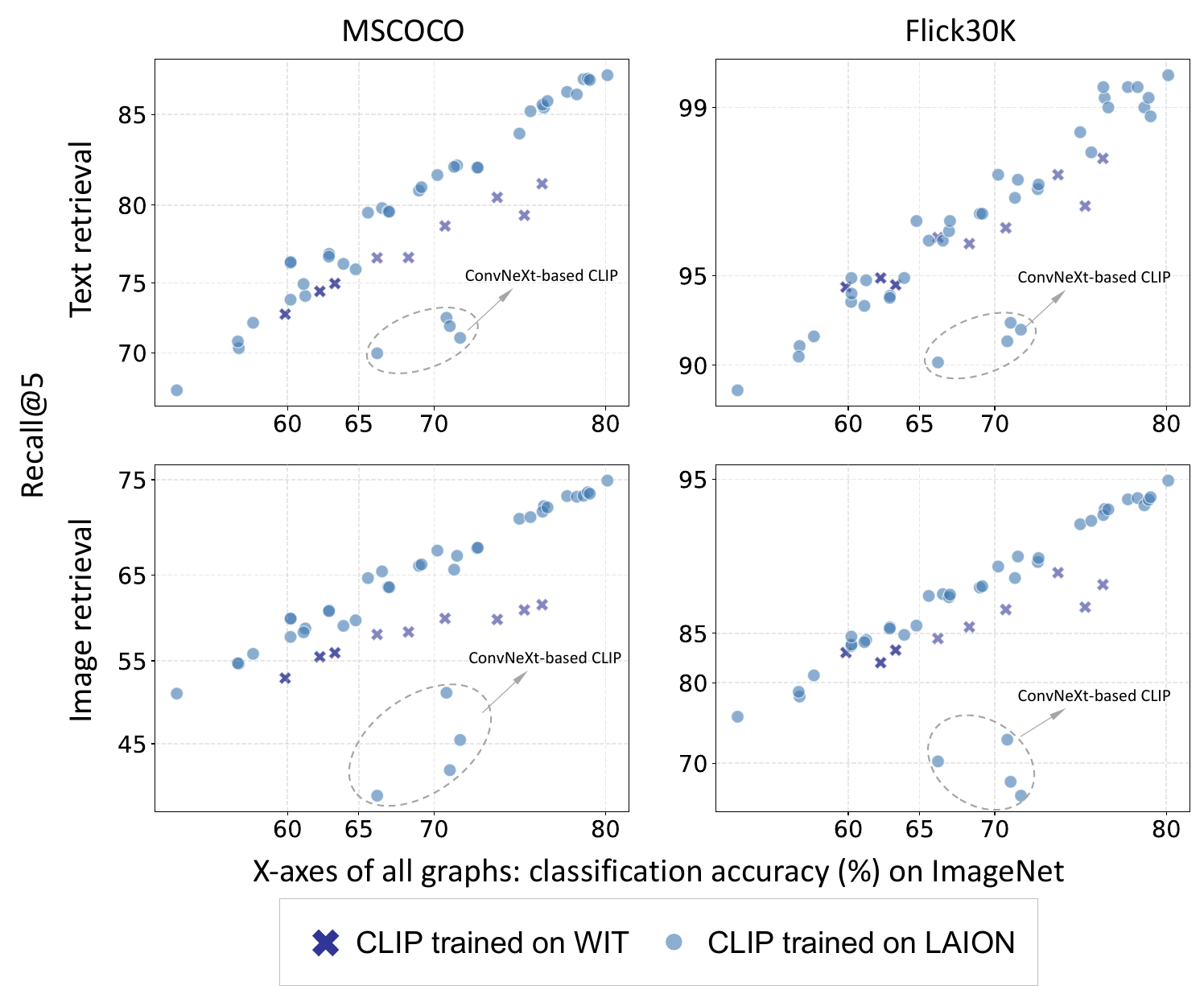}
    \vskip 0.1in
    \caption{
    \textbf{Image/text zero-shot retrieval \emph{v.s} classification accuracy} on MSCOCO and Flick30K measured by Recall$@5$. Classification accuracy is predictive of zero-shot retrieval capability. Moreover, four ConvNeXt-based CLIP models trained with a limited range of random resize crop exhibit much lower retrieval performance.
    }
    \label{fig:retrieval}
    \vskip 0.1in
\end{figure}

\section{Zero-Shot Retrieval} \label{sec:retrieval}
Since CLIP models are trained using contrastive loss to associate text and image pairs, we evaluate their zero-shot retrieval capability on the Flickr30K~\cite{young-etal-2014-image} and MSCOCO~\cite{chen2015microsoft} datasets in this section.

We have three major observations on the two datasets. \textbf{First}, CLIP's zero-shot retrieval capability correlates with its image classification performance. Fig.~\ref{fig:retrieval} illustrates image and text zero-shot retrieval (gauged by Recall$@5$) against their accuracy on ImageNet. We observe that classification ability is predictive of their retrieval capability. 
\textbf{Second}, training distribution deviates from the retrieval performance trend. Specifically, CLIP models trained on WIT slightly deviate from the trend formed by CLIP models trained on LAION, and the training quantity does not affect the trend.
\wj{\textbf{Last}, we observe four specific ConvNeXt-based CLIP models significantly depart from the trend of LAION. We notice that they are trained with a limited random resize crop range ($0.9, 1.0$). This limited augmentation likely reduces training view diversity, resulting in less variation in object scale and context. In image-text retrieval, where the model must extract consistent global representations that align well with corresponding textual descriptions, this lack of variation can hinder the learning of robust embeddings, ultimately affecting retrieval performance.}
While this work does not consider such training augmentations, it would be interesting to explore their impact on retrieval.

\section{3D Awareness}\label{sec:3d}
CLIP models are trained using contrastive loss to associate text and image pairs in feature space, but this training does not explicitly incorporate 3D understanding, such as recognizing geometric concepts like multi-view consistency and depth. Despite being trained on 2D data, recent studies suggest that models like CLIP can still be effective in 3D-related tasks~\mbox{\cite{el2024probing, zhang2024telling, zhang2024tale}}. Building on this insight, this section evaluates the behaviors of CLIP models in 3D-specific scenarios, particularly examining their ability to capture 3D geometry and their robustness to 3D distortions.

\begin{figure*}[!ht]
    \centering
    \includegraphics[width=1\linewidth]{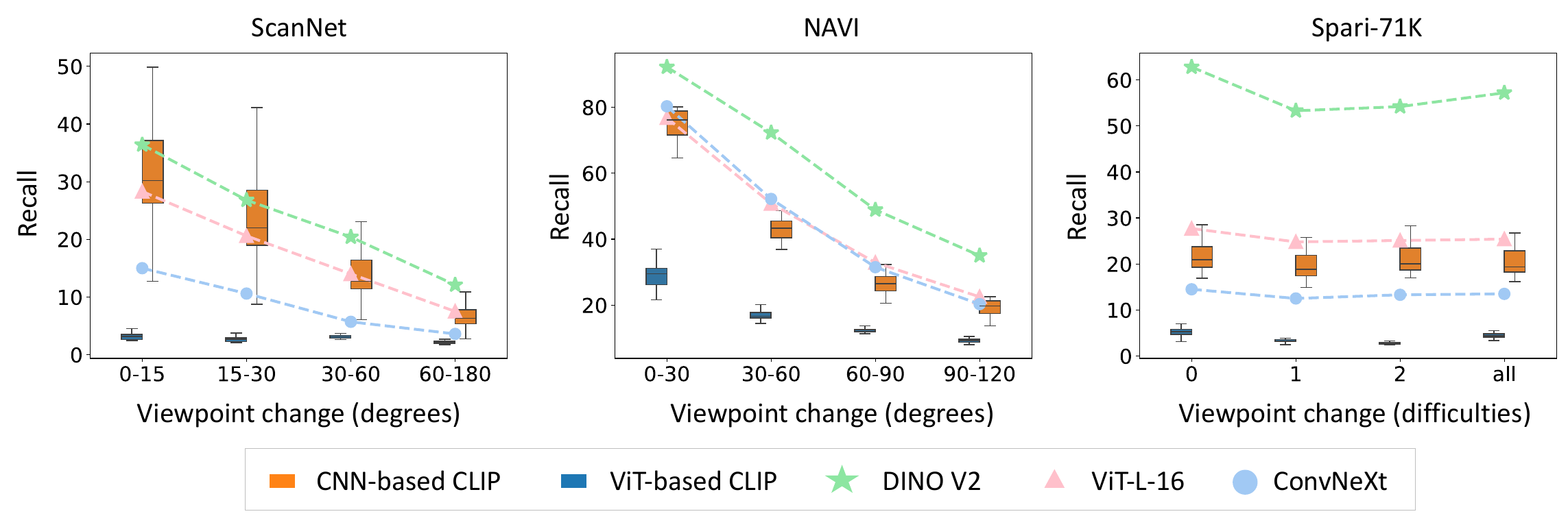}
    \caption{
     \textbf{Correspondence matching performance (Recall $\uparrow$) with respect to their viewpoint change}. We report results on geometric correspondence matching (ScanNet, NAVI) and semantic correspondence matching (Spair-71K). CLIP models are grouped by the architecture of the visual encoder into CNN-based and ViT-based.
    We observe that CNN-based CLIP models consistently outperform ViT-based CLIP models, particularly in scenarios with larger viewpoint variations, and achieve competitive results compared to supervised models like ConvNeXt and ViT-L-16.
    }
    \label{fig:corre}
\end{figure*}

\subsection{Correspondence Matching}

\noindent \textbf{Geometric Correspondence.} Given two views of the same object or scene, the objective is to identify pixels in both views that correspond to the same location in 3D space. We evaluate this using recall on the ScanNet~\cite{dai2017scannet} dataset for object-centric correspondence and NAVI~\cite{jampani2023navi} for scene-centric correspondence. Correspondence recall measures the percentage of correct correspondences that fall within a defined threshold distance. Following the protocol in~\cite{el2024probing}, we categorize performance based on the magnitude of transformation between view pairs.

\noindent \textbf{Semantic Correspondence.} This task generalizes geometric correspondence by requiring matching of semantically similar parts across different instances of the same object class. For example, mapping the left paw of two different dogs. We use the SPair-71K~\cite{min2019spair} dataset, with performance measured by recall. Similar to geometric correspondence, we group results by the degree of view variation. Fig.\ref{fig:corre} groups CLIP models  based on their visual encoder architectures (CNN-based and ViT-based). For comparison, we also include standard supervised models such as ConvNeXt and ViT-L/16 (DeiT III)\cite{touvron2022deit}, which are trained on ImageNet-22K, alongside DINO-V2~\cite{oquab2023dinov2}.

\vspace{0.2cm}
\noindent \textbf{Observations.} 
First, ViT-based CLIP models exhibit weaker performance across three datasets (ScanNet, NAVI, and Spari-71K), falling behind the supervised model (ViT-L-16), which also uses a transformer-based architecture. In contrast, CNN-based CLIPs consistently achieve higher recall scores than their ViT-based counterparts, particularly as viewpoint changes become more extreme. Additionally, CNN-based CLIP models show competitive performance when compared to supervised CNN model ConvNeXt.
This suggests the combined effect of the visual encoder architecture and training objective, which plays a crucial role in influencing CLIP’s ability to manage correspondence matching.
Second, our study extends the observation of~\cite{el2024probing}, showing CNN-based CLIP models not only perform competitively with ViT-L/16 on NAVI but also match DINO-V2 on ScanNet. Note that, DINO-V2 emerges as the top performer across all three datasets.
These findings suggest that CNN-based CLIPs generally exhibit stronger correspondence matching than ViT-based CLIPs, especially in scenarios involving significant viewpoint variations.

\begin{figure*}[t]
    \centering
    \includegraphics[width=1\linewidth]{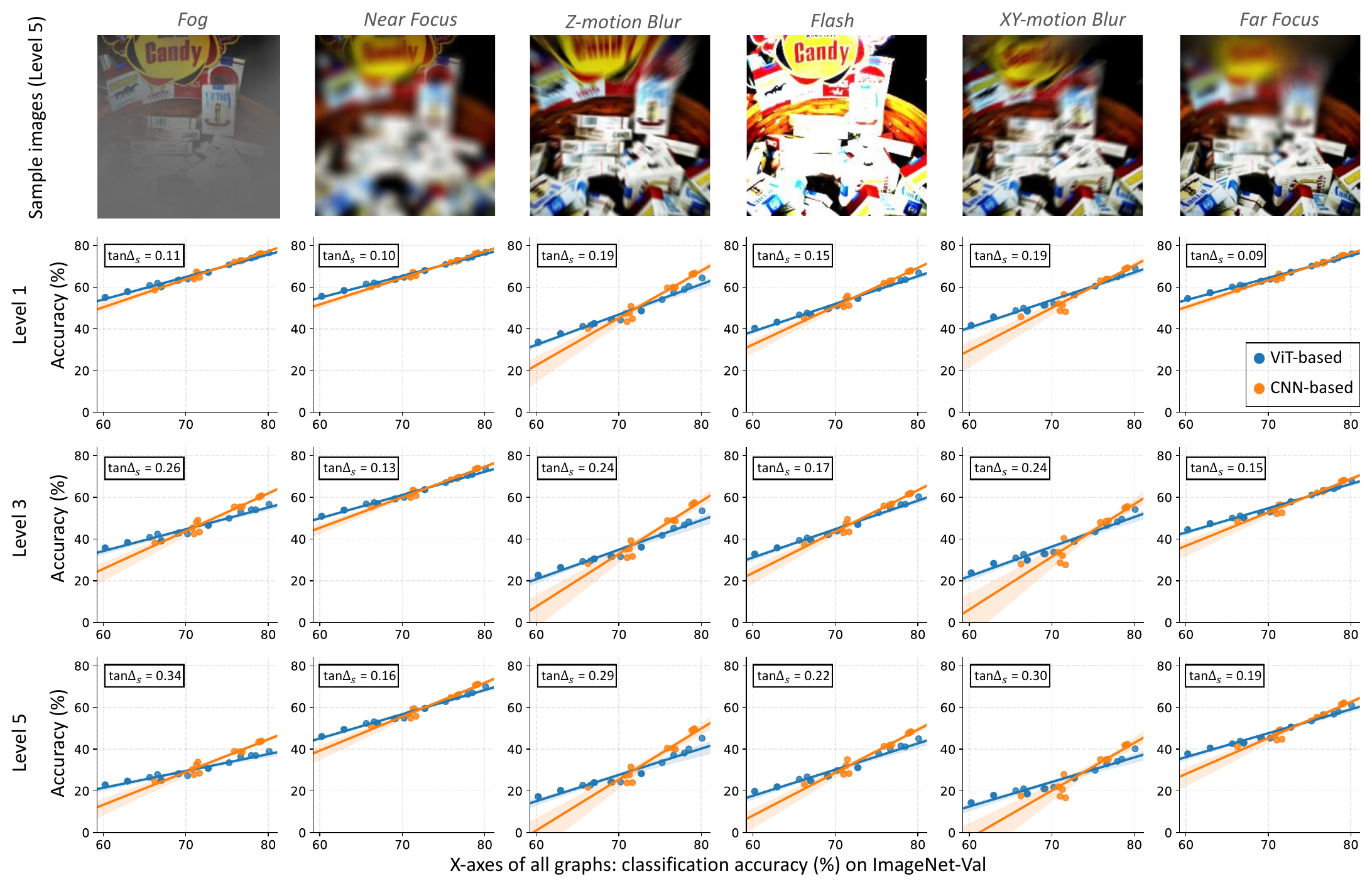}
    \caption{\textbf{Robustness comparison of ViT-based and CNN-based CLIP models under varying 3D-related corruptions.} The x-axis represents accuracy on ImageNet-Val, while the y-axis represents accuracy on the corrupted dataset.
    We show the accuracy of ViT-based and CNN-based CLIP models across six types of 3D-related corruptions: \textit{Fog}, \textit{Near Focus}, \textit{Z-motion Blur}, \textit{Flash}, \textit{XY-motion Blur}, and \textit{Far Focus}, evaluated at three severity levels (Level $1$, Level $3$, and Level $5$). Each column shows that CNN-based models consistently exhibit steeper slopes, indicating greater resilience with less performance degradation as ImageNet-Val accuracy improves. As corruption intensity increases, the gap between the slopes, represented by $tan(\Delta_S)$, widens, particularly under severe conditions like \textit{Fog} and \textit{Z-motion Blur}. This widening gap highlights the superior robustness of CNN-based models compared to their ViT-based counterparts, especially at higher corruption levels. This reinforces the significant impact of visual encoder architecture on CLIP's ability to handle 3D-related corruption. Sample images of Level $5$ severity for each corruption are provided on the top for reference.}
    \label{fig:3dcc}
\end{figure*}

\subsection{Robustness against 3D corruptions}
We further evaluate the ability of CLIP models to handle 3D-related corruptions using the 3D Common Corruptions (3DCC) benchmark~\cite{kar20223d}, which applies corruptions based on 3D transformations. Unlike the common corruptions in~\cite{hendrycks2019benchmarking}, these transformations consider the underlying geometry of the scene, producing distortions that are more reflective of real-world conditions. Sample images of corruptions are shown in the last row in Fig.~\ref{fig:3dcc}. For example, the \textit{fog} gets denser further away from the camera.
In this study, we analyze six types of 3D-related corruptions, each with five severity levels, and examine only CLIP models pre-trained on LAION to maintain consistency in training dataset distributions. Based on correspondence matching, we categorize the CLIP models into CNN-based and ViT-based groups.

\vspace{0.2cm}
\noindent\textbf{CNN-based CLIP models demonstrate stronger robustness to 3D-related corruptions as corruption intensity increases.} Fig.~\ref{fig:3dcc} shows the performance of ViT-based and CNN-based CLIP models across various 3D-related corruptions (\textit{Fog}, \textit{Near Focus}, \textit{Z-motion Blur}, \textit{Flash}, \textit{XY-motion-blur} and \textit{Flash}) at different severity levels (Level $1$, Level $3$, and Level $5$). For each row, the slope of the CNN-based models is consistently steeper than that of the ViT-based models, indicating that CNN-based models experience less degradation in performance as the clean ImageNet validation accuracy increases. This suggests that CNN-based models are more robust in maintaining accuracy under 3D distortions.

Furthermore, as the corruption intensity increases (moving from Level 1 to Level 5), the gap between the slopes, represented by $tan(\Delta_S)$, widens. This increase highlights that the advantage of CNN-based models becomes more pronounced under higher severity of corruptions, particularly for challenging distortions like \textit{Fog} and \textit{Z-motion Blur}. The growing slope difference indicates that CNN-based models are increasingly more capable of handling severe 3D corruptions compared to ViT-based models. These results reinforce the importance of visual encoder architecture in achieving robustness across varying corruption intensities, with CNN-based models consistently outperforming ViT-based models, especially as the corruption severity escalates.
When considered alongside the results from the correspondence matching, these findings underscore the pivotal role that visual encoder architecture plays in enhancing robustness to 3D corruptions.
\wj{Lastly, ViT-based CLIP models struggle with 3D geometric understanding, whereas DINO models perform better. This has implications for downstream multimodal models like LLaVA~\cite{liu2024visual}, which typically rely on CLIP backbones. Combining DINO with CNN-based CLIP features could improve spatial reasoning, as suggested in recent study~\cite{kar2024brave}.}

\section{Visual and Language Encoder Interaction: A Classification Perspective} \label{sec:llava}
Modern large multimodal models (LMMs), such as LLaVA~\cite{liu2024visual}, typically use a frozen pre-trained visual encoder from CLIP as their visual backbone, with instruction fine-tuning applied to the linear projector and the language model components. This raises an important question: how does the interaction between a shared visual encoder and distinct language models affect the classification performance of LLaVA compared to CLIP-like models?

Driven by this, we compare the classification accuracy of CLIP and LLaVA to investigate how the interaction between the shared visual encoder and their distinct language models influences overall performance. In this section, ``LLaVA" and ``CLIP" refer to their training paradigms rather than specific model implementations. We also include SigLIP~\cite{zhai2023sigmoid} as another representative of CLIP-like models.

Our evaluation is conducted on three splits of the \mbox{ImageNet-D} dataset~\cite{zhang2024imagenet}: \textit{Background}, \textit{Texture}, and \textit{Material}. This dataset, generated by a text-to-image diffusion model, poses significant classification challenges. We adopt a VQA-style approach for LLaVA’s classification, providing it with a category list per image and prompting it to select the correct category. The list includes the ground truth (GT) category and three ``failure" categories—incorrect categories ranked with the highest confidence by a pretrained category selection model—ensuring a unique category list for each image. We evaluate the role of the category selection model using \mbox{ResNet-50}, \mbox{CLIP-ViT-L/14-336}, and \mbox{SigLIP-SO-14}. 

To explore the interaction between the language and CLIP vision encoders, we consider six LLaVA models, combining two types of visual encoders—CLIP-\mbox{ViT-L/14-336} and \mbox{SigLIP-SO-14}—and three language encoders: \mbox{Mistral-Instruct-V2}~\cite{jiang2023mistral}, Llama2-Chat~\cite{touvron2023llama}, and Vicuna-V2-7B~\cite{touvron2023llama}. 
For a fair comparison, CLIP is given the same category list using the default prompt template by~\cite{radford2021learning} (e.g., ``a photo of [category]"). LLaVA’s prompt format is:

\begin{tcolorbox}

\texttt{What is the main object in this image? Choose from the following list:}
    
\texttt{A.[Ground truth class]}

\texttt{B.[Failure class 1]}

\texttt{C.[Failure class 2]}

\texttt{D.[Failure class 3]}

\texttt{Please answer the question using the choice from the list.}
\end{tcolorbox}

\begin{table}
    \centering
    \Large
    \begin{adjustbox}{width=\linewidth}
    \begin{tabular}{c c c c c ccc}
    \toprule    
     No. & Category List & Visual encoder & Type & LLM &  Background & Material & Texture \\ 
    \midrule

    \multirow{8}{*}{\makecell{1}} & \multirow{8}{*}{\makecell{ ResNet-50}} & 
    \multicolumn{1}{c}{\multirow{4}{*}{\makecell{CLIP \\ ViT-L/14-336 \\ (WIT)}}} 
    & CLIP & {-} & {$0.90$} & $0.92$ & $0.95$\\
   
   & & & LLaVA & {Mistral-Instruct-V2} & {$0.82$} & $0.81$ & $0.80$\\
   
   & & & LLaVA & {Llama2-Chat} & {$0.91$} & $0.87$ & $0.86$\\
   
   & & & LLaVA & {Vicuna-V2-7B} & $0.92$ & $0.89$ & $0.91$ \\
    
    \cmidrule(lr{1em}){3-8}
    
   & & \multicolumn{1}{c}{\multirow{4}{*}{\makecell{SigLIP-SO-14}}} 
    & CLIP & {-} & {$0.97$} & $0.96$ & $0.99$\\
   
   & & & LLaVA  & {Mistral-Instruct-V2} & {$0.84$} &$0.73$& $0.79$\\
   
   & & & LLaVA & {Llama2-Chat} & {$0.93$} &$0.91$& $0.93$\\
   
   & & & LLaVA & {Vicuna-V2-7B} & $0.92$ & $0.90$ & $0.94$ \\

    \midrule

    \multirow{8}{*}{\makecell{2}}  & \multirow{8}{*}{\makecell{CLIP \\ SigLIP-SO-14 \\ (Webli)}} & 
    \multicolumn{1}{c}{\multirow{4}{*}{\makecell{ViT-L/14-336 \\ (WIT)}}} 
    & CLIP & {-} & {$0.23$} & $0.24$ & $0.21$\\
    
    & & & LLaVA & {Mistral-Instruct-V2} & {$0.41$} & $0.35$ & $0.28$\\
   
   & & & LLaVA & {Llama2-Chat} & {$0.52$} & $0.35$ & $0.34$\\
   
   & & & LLaVA & {Vicuna-V2-7B} & $0.57$ & $0.48$ & $0.42$ \\

    \cmidrule(lr{1em}){3-8}
    
   & & \multicolumn{1}{c}{\multirow{4}{*}{\makecell{SigLIP-SO-14}}} 
    & CLIP & {-} & {$0.65$} & $0.61$ & $0.61$\\
    
    & & & LLaVA  & {Mistral-Instruct-V2} & {$0.44$} &$0.33$& $0.35$\\
   
   & & & LLaVA & {Llama2-Chat} & {$0.60$} &$0.47$& $0.46$\\
   
   & & & LLaVA & {Vicuna-V2-7B} & $0.59$ & $0.48$ & $0.43$ \\

   \midrule

   \multirow{8}{*}{\makecell{3}} & \multirow{8}{*}{\makecell{CLIP \\ ViT-L/14-336 \\ (WIT)}} & 
    \multicolumn{1}{c}{\multirow{4}{*}{\makecell{ViT-L/14-336 \\ (WIT)}}} 
    & CLIP & {-} & {$0.14$} & $0.14$ & $0.13$\\
    
    & & & LLaVA & {Mistral-Instruct-V2} & {$0.37$} & $0.35$ & $0.25$\\
   
   & & & LLaVA & {Llama2-Chat} & {$0.49$} & $0.32$ & $0.30$\\
   
   & & & LLaVA & {Vicuna-V2-7B} & $0.57$ & $0.45$ & $0.42$ \\

    \cmidrule(lr{1em}){3-8}
    
   & & \multicolumn{1}{c}{\multirow{4}{*}{\makecell{SigLIP-SO-14}}} 
    & CLIP & {-} & {$0.69$} & $0.67$ & $0.65$\\

   & & & LLaVA  & {Mistral-Instruct-V2} & {$0.46$} &$0.34$& $0.36$\\
   
   & & & LLaVA & {Llama2-Chat} & {$0.62$} &$0.48$& $0.48$\\
   
   & & & LLaVA & {Vicuna-V2-7B} & $0.59$ & $0.52$ & $0.44$ \\ 
    
    \bottomrule 
    \end{tabular}
    \end{adjustbox}
    \caption{\textbf{Compared CLIP and LLaVA models on ImageNet-D.} We include two visual backbones: CLIP-L/14-336 and SigLIP-SO-L and two language models for LLaVA: Mistral-Instruct-V2, Llama2-Chat, and Vicuna-V2-7B.}
    \label{tab:llava}
\end{table}

\vspace{0.15cm}
\noindent\textbf{Observations}: 
We report the results on \mbox{ImageNet-D} in Table \ref{tab:llava} and summarize the observations as follows.

\textbf{First}, extending the findings of~\cite{zhang2024imagenet}, which evaluate CLIP (ViT/14) as a category selection model, we find that the interactions between vision and language components in selection-based networks vary significantly with task difficulty.
When the category list is easy for CLIP, LLaVA models using the same visual encoder do not yield consistent improvements and sometimes exhibit slight performance drops. In contrast, when the category list is challenging for CLIP, LLaVA models using the same visual encoder show substantial gains. For example, in row 1, the most confused categories of ResNet-50 are easy for CLIP, and LLaVA brings no improvement. Similarly, in row 2, when SigLIP-SO-14 performs well, LLaVA shows a performance drop. However, in the same row, when the category list becomes difficult for SigLIP-SO-14, LLaVA improves accuracy by over 20\% across three splits. A similar pattern is observed in row 3 for CLIP (ViT-L/14-336) and its LLaVA counterpart.
\wj{Since LLaVA and CLIP share the same visual encoder, we speculate that the observed gains arise when CLIP’s visual-text alignment is weak—likely due to ambiguous categories or limited pre-training coverage. In such cases, LLaVA’s language model may help disambiguate visual features by leveraging external knowledge acquired during multimodal instruction tuning. Conversely, when CLIP already handles the token comparison effectively, the language model may over-interpret the visual input, occasionally leading to reduced performance.}

\textbf{Second}, the choice of language model (LLM) within LLaVA significantly impacts classification performance. Vicuna-V2-7B consistently outperforms Mistral-Instruct-V2, while the choice of visual encoder also plays a critical role. LLaVA models built on SigLIP-SO-14 outperform those using ViT-L/14-336, echoing recent findings in the literature.

\wj{The above suggests that LLaVA’s performance gains are not solely the result of architectural complexity or additional training data, but rather the interaction between the two. The model’s effectiveness depends on how the language model, visual encoder, and learned projection layer work together in response to varying input complexity. This indicates the importance of designing vision-language models with careful attention to how components are integrated and how they respond under different levels of visual-text alignment difficulty.}

\section{Impact of Training and Inference Strategy on Model Robustness}

\begin{figure*}[t!]
\begin{minipage}{0.5\linewidth}
    \label{fig:ablation}
    \includegraphics[width=\linewidth]{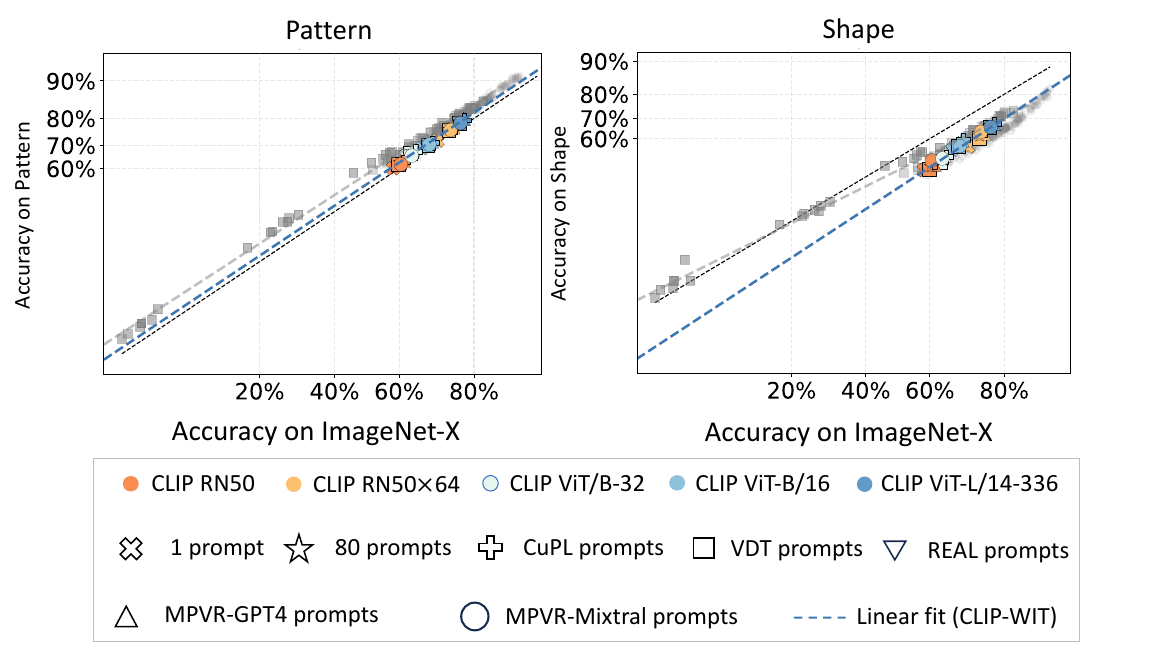}
    \figcaption{\wj{\textbf{Influence of test-time prompts on CLIP’s visual-factor robustness}. We evaluate five CLIP models trained on WIT, represented by different colors for architectures and different shapes. The dashed grey line represents robust linear regression~\cite{huber2011robust} based on the original CLIP-WIT models with $80$ prompts. Different prompt sets may influence classification performance but do not significantly impact visual factor robustness because models still lie on the original line.}}
     \label{fig:prompt}
\end{minipage}
\:
\begin{minipage}{0.48\linewidth}
\centering
\begin{adjustbox}{width=\linewidth}
    \begin{tabular}{c c c c c c c}
\toprule
\multirow{3}{*}{\textbf{Backbone}} & \multirow{3}{*}{\textbf{Pre-training dataset}} & \multirow{3}{*}{\textbf{Classification}} & \multicolumn{2}{c}{\textbf{OOD Detection}} & \multicolumn{2}{c}{\textbf{Calibration}} \\
\cmidrule(lr){4-5}  \cmidrule(lr){6-7}
&&&&&Before-temp & After-temp \\
&&Accuracy ($\uparrow$) & AUROC ($\uparrow$) & FPR ($\downarrow$) & ECE ($\downarrow$) & ECE ($\downarrow$) \\
\midrule
\multirow{7}{*}{\textbf{RN50}}  & 1 Pompt & $0.41$ & \colorbox{green(munsell)!20}{$0.84$} & \colorbox{green(munsell)!20}{$0.61$} & $0.08$ &  $0.08$  \\  
          & 80 Prompts & $0.43$       & $0.83$ & $0.64$ & $0.08$ & \colorbox{green(munsell)!20}{$0.08$}  \\  
          & CuPL & \colorbox{green(munsell)!20}{$0.44$}            & $0.83$ & $0.63$ & $0.09$ & $0.09$ \\  
          & VDT        & $0.42$       & $0.82$ & $0.65$ & $0.10$ & $0.09$  \\
          & MPVR-GPT4 & {$0.43$} & $0.82$ & $0.65$ & $0.09$ &$0.09$  \\
          & MPVR-Mistral & $0.43$ & $0.82$ & $0.66$ & \colorbox{green(munsell)!20}{$0.08$} & $0.09$ \\ 
          & REAL  & $0.42$       & $0.81$ & $0.69$ & $0.08$ & $0.09$  \\
\midrule
\multirow{7}{*}{\textbf{ViT-B/16}}  & 1 Pompt  & $0.57$         & $0.86$ & $0.55$ & $0.05$ & $0.06$  \\  
          & 80 Prompts  & $0.59$      & $0.85$ & $0.57$ & $0.05$ & $0.06$  \\  
          & CuPL   & $0.60$          & $0.86$ & $0.54$ & $0.06$ & $0.06$ \\  
          & VDT  & $0.59$             & \colorbox{green(munsell)!20}{$0.86$} & \colorbox{green(munsell)!20}{$0.54$} & $0.07$ & $0.06$  \\ 
          & MPVR-GPT4 & \colorbox{green(munsell)!20}{$0.60$} & $0.85$ & $0.58$ & $0.05$ &$0.06$  \\
          & MPVR-Mistral & {$0.60$} & $0.85$ & $0.60$ & \colorbox{green(munsell)!20}{$0.05$} & \colorbox{green(munsell)!20}{$0.06$}  \\  
          & REAL  & {$0.58$}      & $0.83$ & $0.63$ & {$0.06$} & {$0.06$}  \\ 
\bottomrule
\end{tabular}
	\end{adjustbox}
	  \vspace{0.4cm}
	\tabcaption{\wj{\textbf{Influence of test-time prompts on CLIP’s classification, OOD detection and calibration}. We evaluate CLIP models trained on WIT with ResNet-50 and ViT-B/16 as the visual encoder. We find that prompt sets generated by large language models may improve zero-shot CLIP models' classification accuracy, but it does not enhance other OOD detection or calibration}.}
\label{tab:prompt}
\end{minipage}
\end{figure*}

\subsection{Impact of Test-Time Prompts} \label{sec:prompt}

In the previous analyses, we used the default prompt set provided by~\cite{radford2021learning}. Here, we investigate how varying test-time prompts influence CLIP's performance in out-of-distribution (OOD) detection, visual factor robustness, and predictive uncertainty. \wj{We experiment with five additional prompt sets: (1) a single prompt (``a photo of a \{label\}''); (2) a set generated by GPT-3 following~\cite{pratt2023does}; (3) a prompt set generated by GPT-4~\cite{achiam2023gpt} using the chain-of-thought strategy~\cite{wei2022chain} (VDT)~\cite{maniparambil2023enhancing}; (4) prompts generated by GPT-4 (MPVR-GPT4) or Mistral (MPVR-Mistral) with a target-task-oriented design~\cite{mirza2024meta}; (5) prompts generated by GPT-4 with Retrieval-Augmented Learning (REAL)~\cite{parashar2024neglected}.} These prompts are tested across five CLIP models—\mbox{RN50}, \mbox{RN50×64}, \mbox{ViT-B/16}, \mbox{ViT-B/32}, and \mbox{ViT-L/14-336}—all trained on the WIT dataset.

\wj{Fig.\ref{fig:prompt} and Table\ref{tab:prompt} summarize the effects of various prompt strategies on CLIP’s classification, robustness, OOD detection, and calibration. Using fewer prompts (\textit{e.g.}, a single prompt) reduces classification accuracy but improves OOD detection and calibration. Factor-level robustness, such as on the Pattern task, remains largely unchanged regardless of prompt type, with models following the original CLIP-WIT trend. Prompt sets generated by large language models—including VDT, MPVR-GPT4, MPVR-Mistral, and REAL—consistently improve classification accuracy, but show no clear advantage in OOD detection or calibration. These results highlight a key challenge: how to design prompt strategies that simultaneously enhance other objectives beyond classification accuracy only.}

\begin{figure*}[!ht]
\centering
\includegraphics[width=\linewidth]{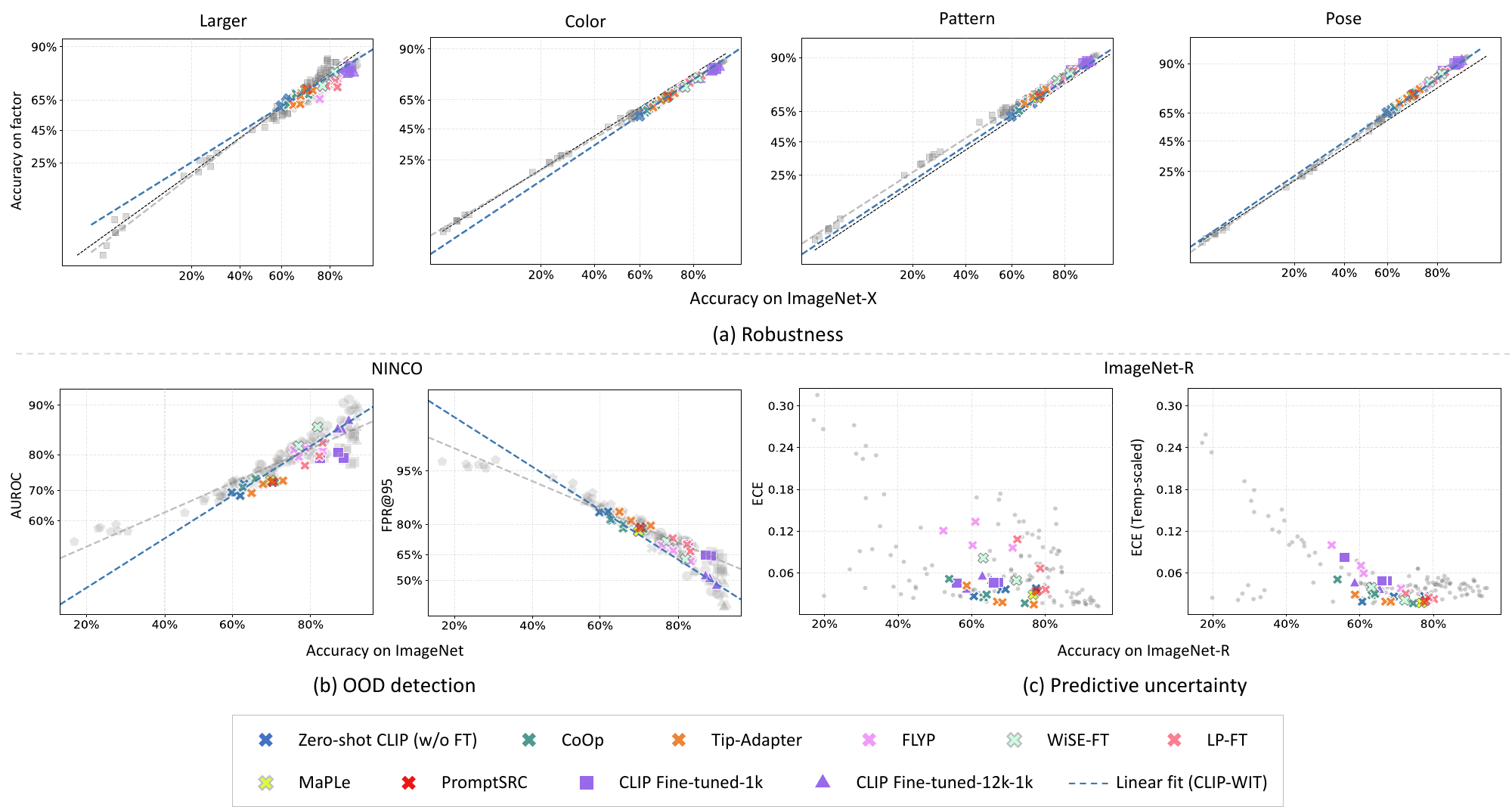}
\vskip 0.05in
\caption{
\textbf{Influence of fine-tuning algorithms on CLIP’s robustness, OOD detection, and predictive uncertainty}. We fine-tune four CLIP models trained on WIT using various algorithms. Different colors represent model architectures, and different shapes denote fine-tuning algorithms. The blue dashed line is fit with robust linear regression~\cite{huber2011robust} for original CLIP-WIT models, while the grey dashed line represents zero-shot CLIP trained on LAION. Results show that contrastive fine-tuning improves overall classification accuracy but negatively impacts predictive uncertainty.
}
\label{fig:peft}
\end{figure*}

\subsection{Effect of Fine-Tuning Procedures} \label{sec:ft}

\wj{In addition to standard fine-tuning methods (\textit{i.e.}, cross-entropy fine-tuning on ImageNet), we examine seven alternative fine-tuning strategies: contrastive fine-tuning (FLYP) as introduced by~\cite{goyal2023finetune}, two robsut fine-tuning methods--WiSE-FT~\cite{wortsman2022robust} and LP-FT~\cite{kumar2022fine}, and four parameter-efficient methods--CoOp~\cite{zhou2022learning}, Tip-Adapter~\cite{zhang2022tip}, MaPLe~\cite{khattakMaPLe} and PromptSRC~\cite{Khattak_2023_ICCV}. They are applied to fine-tune zero-shot CLIP models pre-trained on WIT.}

\wj{In Fig.~\ref{fig:peft}, we present the performance of fine-tuned CLIP models across visual factor robustness, OOD detection, and calibration. The results reveal mixed effects across different tuning methods. For visual factor robustness, CoOp and PromptSRC preserve the properties of zero-shot CLIP, aligning with prior findings that test-time prompts have limited influence on robustness. FLYP and Tip-Adapter improve robustness against the \textit{Pattern} factor but reduce it under \textit{Larger} visual changes. LP-FT maintains robustness across several factors, while WiSE-FT slightly weakens it on \textit{Larger}. On OOD detection, most methods---including FLYP, LP-FT, WiSE-FT, and PromptSRC---enhances both accuracy and detection performance, with LP-FT showing strong generalization as their models lie above the zero-shot CLIP-WIT trend. For calibration, FLYP increases calibration error, while CoOp, Tip-Adapter, and PromptSRC maintain well-calibrated predictions. LP-FT and WiSE-FT increase error slightly before temperature scaling but recover calibration performance afterward, outperforming FLYP in uncertainty estimation.
}

These findings suggest that while fine-tuning can improve certain aspects of CLIP’s performance, achieving a balance between classification accuracy, OOD detection, and predictive uncertainty remains a challenge, highlighting the need for further research into fine-tuning strategies that can address all of these objectives.

\subsection{Extending to Different Training Paradigms} \label{sec:paradigm}
\noindent\wj{We have expanded our analysis beyond CLIP to include four additional vision-language training paradigms: (1) BLIP~\cite{goyal2023finetune}, (2) BLIP-2~\cite{li2023blip}, (3) SigLIP~\cite{zhai2023sigmoid}, and (4) ViTamin~\cite{chen2024vitamin}.
We summarized the results in Fig.~\ref{fig:paradigm} and Table~\ref{tab:paradigm}
We observe that different training paradigms yield trade-offs across robustness, OOD detection, calibration, and 3D performance. SigLIP achieves the best overall balance, with strong OOD detection, low calibration error, and competitive 3D robustness.\mbox{ViTamin-L-256px} leads in classification accuracy and 3D robustness but suffers from poor calibration. In addition, BLIP and BLIP-2 consistently underperform across most metrics. These results reinforce our main claim: no single paradigm excels universally, underscoring the need for multi-dimensional evaluation beyond accuracy alone.}

\begin{figure*}[t!]
\begin{minipage}{0.5\linewidth}
    \label{fig:ablation}
    \includegraphics[width=\linewidth]{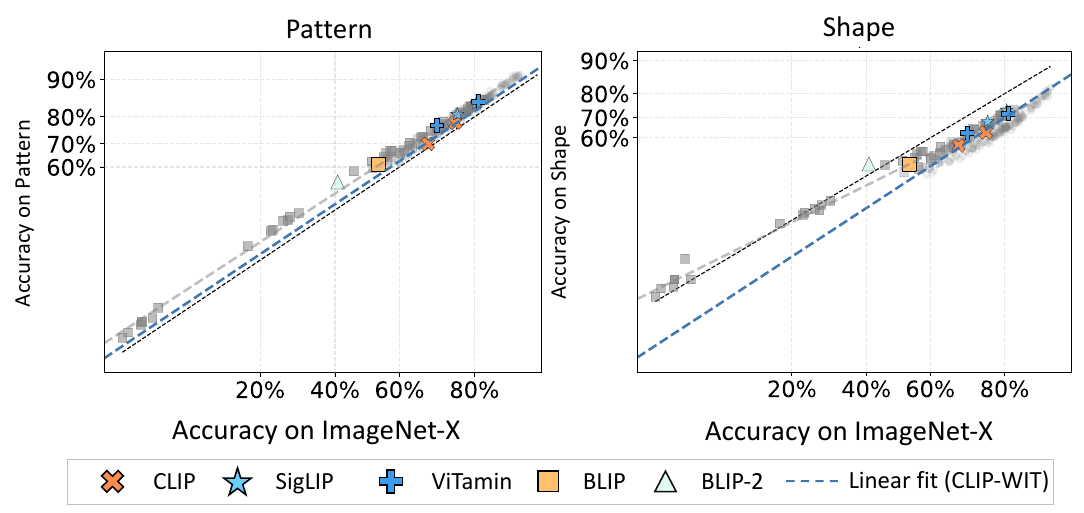}
    \figcaption{\wj{\textbf{Visual-factor robustness of different training paradigms from CLIP}. We evaluate five vision-language training paradigms: CLIP, BLIP, BLIP-2, SigLIP and ViTamin. The dashed blue line represents robust linear regression~\cite{huber2011robust} based on the original CLIP-WIT models. Different training paradigms effectively impact visual factor robustness.}}
     \label{fig:paradigm}
\end{minipage}
\:
\begin{minipage}{0.48\linewidth}
\centering
\begin{adjustbox}{width=\linewidth}
    \begin{tabular}{c c c c c c c}
\toprule
\multirow{3}{*}{\textbf{Training paradigm}} & \multirow{3}{*}{\textbf{Classification}} & \multicolumn{2}{c}{\textbf{OOD Detection}} & \multicolumn{2}{c}{\textbf{Calibration}} & \multirow{3}{*}{\textbf{3D Robustness}}\\
\cmidrule(lr){3-4}  \cmidrule(lr){5-6}
&&&&Before-temp & After-temp \\
&Accuracy ($\uparrow$) & AUROC ($\uparrow$) & FPR ($\downarrow$) & ECE ($\downarrow$) & ECE ($\downarrow$)&Accuracy ($\uparrow$) \\
\midrule
 CLIP-ViT-B/16 & $0.59$ &{$0.85$} & {$0.57$} & \colorbox{green(munsell)!20}{$0.06$} &  $0.06$  & $0.34$\\  
 BLIP-Base & $0.51$       & $0.76$ & $0.70$ & $0.07$ & {$0.08$} & $0.27$ \\  
 SigLIP-ViT-B/16 & \colorbox{green(munsell)!20}{$0.68$}            & \colorbox{green(munsell)!20}{$0.89$} & \colorbox{green(munsell)!20}{$0.50$} & $0.08$ & \colorbox{green(munsell)!20}{$0.05$} & \colorbox{green(munsell)!20}{$0.38$}\\  
 ViTamin-Base        & $0.62$       & $0.84$ & $0.67$ & $0.37$ & $0.23$  & $0.35$\\
\midrule
CLIP-ViT-L/14 & $0.72$ & {$0.88$} & {$0.47$} & \colorbox{green(munsell)!20}{$0.06$} &  $0.05$  & $0.46$\\  
 BLIP-2 & $0.53$       & $0.66$ & $0.89$ & $0.14$ & {$0.08$}  & $0.37$\\  
 SigLIP-ViT-L/16 & {$0.75$}            & \colorbox{green(munsell)!20}{$0.91$} & \colorbox{green(munsell)!20}{$0.40$} & $0.07$ & \colorbox{green(munsell)!20}{$0.04$} & $0.46$ \\  
 ViTamin-L-256px        & \colorbox{green(munsell)!20}{$0.79$}       & $0.89$ & $0.49$ & $0.21$ & $0.17$  & \colorbox{green(munsell)!20}{$0.53$}\\
\bottomrule
\end{tabular}
	\end{adjustbox}
	  \vspace{0.6cm}
	\tabcaption{\wj{\textbf{Performance of various training paradigms on classification, OOD detection, calibration and 3D robustness}. We evaluate five vision-language training paradigms: CLIP, BLIP, BLIP-2, SigLIP and ViTamin. We find that no training paradigm is the most performant on all considered safety-related objectives.}}
\label{tab:paradigm}
\end{minipage}
\end{figure*}

\begin{table*}[ht]
\centering
\large
\caption{\textbf{Comparison of CLIP trained with filtered pre-training data on six tasks.} For the classification task, we report average accuracy on ImageNet validation, ImageNet-V2-A, ImageNet-S, ObjectNet, ImageNet-A, ImageNet-R and ImageNet-Vid.
We report averaged AUROC and FPR on NINCO, iNaturalist, DTD, Place, SUN and ImageNet-O. We report ECE before and after calibration. The calibration set is ID-val and test set is the same as OOD generalization. For visual factor robustness, we evaluate \textit{Larger}, \textit{Shape} and \textit{Color}. We use averaged recall@5 to measure text-to-image and image-to-text retrieval on MSCoCo and Flick30K. For 3D robustness, we use accuracy to metric their mean performance on six 3D-related corruptions with severity level 5. The best performance for each architecture is in \colorbox{green(munsell)!20}{green}. We find that data curation technique is an effective method for enhancing model performance beyond classification.}
 \vspace{0.1in}
\renewcommand{\arraystretch}{1.2} 
\begin{adjustbox}{width=\linewidth} %
\begin{tabular}{c c c c c c c c c c c c c}
\toprule
\multirow{3}{*}{\textbf{Backbone}} & \multirow{3}{*}{\textbf{Pre-training dataset}} & \multirow{3}{*}{\textbf{Data Filtering}} & \multirow{3}{*}{\textbf{Classification}}  & \multicolumn{2}{c}{\textbf{OOD Detection}} & \multicolumn{2}{c}{\textbf{Calibration}} & \multicolumn{3}{c}{\textbf{Visual factor robustness}} & \multirow{3}{*}{\textbf{Retrieval}} & \multirow{3}{*}{\textbf{3D robustness}}\\
\cmidrule(lr){5-6}  \cmidrule(lr){7-8}  \cmidrule(lr){9-11}  
    && && & & Before-temp & After-temp & Larger & Shape & Color & &  \\
    && & Accuracy ($\uparrow$) & AUROC ($\uparrow$) & FPR ($\downarrow$) & ECE ($\downarrow$) & ECE ($\downarrow$) & Accuracy ($\uparrow$) & Accuracy ($\uparrow$) & Accuracy ($\uparrow$) & Recall@5 ($\uparrow$) & Accuracy ($\uparrow$)\\
\midrule
\multirow{6}{*}{ViT-B/16} 
          & LAION-400M                & No  & $0.61$ & $0.84$ & $0.65$ & $0.13$ & $\colorbox{green(munsell)!20}{0.05}$ & $0.67$ & $0.56$ & $0.63$ & $0.82$ & $0.32$ \\  
          & MetaCLIP-400M     & Yes & \colorbox{green(munsell)!20}{${0.67}$} & \colorbox{green(munsell)!20}{$0.85$} & \colorbox{green(munsell)!20}{${0.62}$} & \colorbox{green(munsell)!20}{$0.09$} & {$0.08$} & \colorbox{green(munsell)!20}{${0.75}$} & \colorbox{green(munsell)!20}{${0.61}$} & \colorbox{green(munsell)!20}{${0.67}$} & \colorbox{green(munsell)!20}{${0.83}$} & \colorbox{green(munsell)!20}{${0.35}$} \\ 
          \cmidrule(lr){2-13}
          & LAION-2B                  & No  & $0.64$ & $0.85$ & $0.64$ & $0.13$ & \colorbox{green(munsell)!20}{$0.05$} & $0.69$ & $0.60$ & $0.67$ & $0.84$ & $0.34$ \\  
          & DFN-2B            & Yes & \colorbox{green(munsell)!20}{${0.70}$} & \colorbox{green(munsell)!20}{${0.88}$} & \colorbox{green(munsell)!20}{${0.52}$} & \colorbox{green(munsell)!20}{$0.12$} & $0.07$ & \colorbox{green(munsell)!20}{${0.80}$} & \colorbox{green(munsell)!20}{${0.66}$} & \colorbox{green(munsell)!20}{${0.73}$} & \colorbox{green(munsell)!20}{${0.85}$} & \colorbox{green(munsell)!20}{${0.40}$} \\ 
          \cmidrule(lr){2-13}
          & CommonPool-L              & No  & $0.43$ & $0.73$ & $0.86$ & \colorbox{green(munsell)!20}{${0.06}$} & \colorbox{green(munsell)!20}{$0.07$} & $0.45$ & $0.46$ & \colorbox{green(munsell)!20}{$0.58$} & $0.64$ & $0.19$ \\  
          & CommonPool-L-\textit{CLIP} & Yes & \colorbox{green(munsell)!20}{${0.53}$} & \colorbox{green(munsell)!20}{$0.77$} & \colorbox{green(munsell)!20}{$0.81$} & $0.11$ & \colorbox{green(munsell)!20}{$0.07$} & \colorbox{green(munsell)!20}{$0.61$} & \colorbox{green(munsell)!20}{$0.53$} & $0.56$ & \colorbox{green(munsell)!20}{$0.72$} & \colorbox{green(munsell)!20}{$0.26$} \\  
          \midrule
          \multirow{6}{*}{ViT-L/14} 
          & LAION-400M                & No  & $0.68$ & $0.86$ & $0.59$ & $0.17$ & \colorbox{green(munsell)!20}{$0.06$} & $0.75$ & $0.64$ & $0.70$ & \colorbox{green(munsell)!20}{$0.85$} & $0.38$ \\  
          & MetaCLIP-400M     & Yes & \colorbox{green(munsell)!20}{${0.76}$} & \colorbox{green(munsell)!20}{${0.89}$} & \colorbox{green(munsell)!20}{${0.50}$} & \colorbox{green(munsell)!20}{$0.09$} & \colorbox{green(munsell)!20}{$0.06$} & \colorbox{green(munsell)!20}{${0.74}$} & \colorbox{green(munsell)!20}{${0.67}$} & \colorbox{green(munsell)!20}{${0.74}$} & \colorbox{green(munsell)!20}{$0.85$} & \colorbox{green(munsell)!20}{${0.45}$} \\ \cmidrule(lr){2-13}
          & LAION-2B                  & No  & $0.72$ & $0.88$ & $0.52$ & $0.11$ & $0.04$ & $0.82$ & $0.66$ & $0.72$ & $0.87$ & $0.42$ \\  
          & DFN-2B            & Yes & \colorbox{green(munsell)!20}{${0.78}$} & \colorbox{green(munsell)!20}{${0.91}$} & \colorbox{green(munsell)!20}{${0.39}$} & \colorbox{green(munsell)!20}{$0.07$} & \colorbox{green(munsell)!20}{$0.04$} & \colorbox{green(munsell)!20}{${0.85}$} & \colorbox{green(munsell)!20}{${0.74}$} & \colorbox{green(munsell)!20}{${0.79}$} & \colorbox{green(munsell)!20}{${0.88}$} & \colorbox{green(munsell)!20}{${0.50}$} \\ \cmidrule(lr){2-13}
          & CommonPool-XL             & No  & $0.72$ & $0.87$ & \colorbox{green(munsell)!20}{$0.54$} & \colorbox{green(munsell)!20}{$0.03$} & $0.04$ & $0.72$ & $0.65$ & $0.70$ & $0.80$ & $0.43$ \\  
          & CommonPool-XL-\textit{CLIP}& Yes & \colorbox{green(munsell)!20}{${0.75}$} & \colorbox{green(munsell)!20}{$0.88$} & \colorbox{green(munsell)!20}{$0.54$} & $0.08$ & \colorbox{green(munsell)!20}{$0.03$} & \colorbox{green(munsell)!20}{$0.74$} & $\colorbox{green(munsell)!20}{0.67}$ & \colorbox{green(munsell)!20}{${0.74}$} & \colorbox{green(munsell)!20}{$0.84$} & \colorbox{green(munsell)!20}{$0.46$} \\  
          \midrule
          \multirow{2}{*}{ConvNeXt-Base}
          & LAION-2B & No & $0.64$ & $0.85$ & $0.64$ &\colorbox{green(munsell)!20}{$0.12$}&\colorbox{green(munsell)!20}{$0.05$}&$0.69$ &$0.59$&\colorbox{green(munsell)!20}{$0.67$}&\colorbox{green(munsell)!20}{$0.72$}&\colorbox{green(munsell)!20}{$0.35$}\\
          & LAION-Aesthetic & Yes & \colorbox{green(munsell)!20}{$0.65$} & \colorbox{green(munsell)!20}{$0.85$} & \colorbox{green(munsell)!20}{$0.63$} &$0.14$&\colorbox{green(munsell)!20}{$0.05$}&\colorbox{green(munsell)!20}{$0.71$} &\colorbox{green(munsell)!20}{$0.62$}&\colorbox{green(munsell)!20}{$0.67$}&$0.68$&$0.33$\\
          \bottomrule
\end{tabular}
\end{adjustbox}
\label{tab:filter}
\end{table*}

\subsection{Robustness Evaluation of Dataset Curation} \label{sec:dc}
High-quality training sets are crucial for developing CLIP models, and as a result, recent research has increasingly emphasized dataset curation (DC) to create these datasets~\mbox{\cite{gadre2023datacomp,fang2023data,xu2023demystifying}}. In this work, we extend the evaluation of DC techniques to robustness-related tasks, including out-of-distribution (OOD) detection, calibration, visual factor-level robustness, and 3D corruption.

To ensure a clear and fair comparison, we control the architecture of the CLIP models and categorize the methods based on their pretraining dataset sources. We consider four DC techniques: 1) CommonPool~\cite{gadre2023datacomp}, which uses a trained CLIP model as a filter; 2) MetaCLIP~\cite{xu2023demystifying}, which leverages metadata for curation and balancing of raw web-sourced data; and 3) DFN-2B~\cite{fang2023data}, which employs a network trained on high-quality datasets for filtering; \wj{4) Aesthetic~\cite{laion2022aesthetics}, which is filtered using perceptual hashing for deduplication and an aesthetic score threshold.}

\wj{Table~\ref{tab:filter} shows that DC techniques consistently improve performance in classification, OOD detection, visual factor robustness, and 3D robustness—particularly for transformer-based models. However, their impact on calibration is limited. 
We also evaluate CNN-based CLIP models trained on LAION-Aesthetic and observe that while dataset filtering improves classification and robustness, it shows limited benefits for retrieval, calibration, and 3D robustness.
These observations suggest that the effectiveness of filtering strategies depends on both model architecture and the specific evaluation objective, emphasizing the need for multi-dimensional assessment beyond classification accuracy.}

\section{Conclusion and Discussion}
Our research contributes to the ongoing discussion regarding the robustness and capabilities of CLIP models, particularly in response to visual factor robustness, OOD detection, the reliability of uncertainty estimation, zero-shot retrieval capabilities, and 3D awareness. To achieve these insights, we performed comprehensive experiments and comparative analyses, systematically evaluating CLIP models against diverse model families. Through an in-depth exploration of critical factors—including training sources, contrastive learning objectives, network architecture, fine-tuning strategies, and test-time prompt variations—our findings provide substantial insights into the unique advantages CLIP models offer.

\noindent \textbf{Discussion on Dataset Overlap.}
Given that CLIP models are pretrained on large-scale web-crawled datasets such as LAION-5B, potential overlap with evaluation benchmarks is a valid concern. Prior work suggests that such overlap is unlikely to significantly affect our findings.  For classification robustness, LAION-5B paper~\cite{gadre2023datacomp} and OpenAI~\cite{radford2021learning} report only isolated cases where overlap impacts performance, and do not view it as a major threat to result validity. For OOD detection, Bitterwolf \textit{et al.}~\cite{bitterwolf2023or} show that overlapping class semantics between pretraining (\textit{e.g.}, IN-21K) and test sets (\textit{e.g.}, NINCO) does not substantially alter detection performance. For calibration, since our evaluation datasets are shared with classification and focus on relative model comparison, any sample-level overlap is unlikely to influence conclusions. Moreover, our emphasis on relative trends, rather than absolute scores, further mitigates this concern.

This work leaves open many interesting and promising directions and we discuss a few. \textbf{First}, we offer an analysis of LLaVA and demonstrate that its large language model can assist in classification where CLIP’s text and visual tokens are misaligned. Future work could explore other modern large vision models (LVMs), such as BLIP-3~\cite{xue2024xgen} and Otter~\cite{zhao2024evaluating}, to deepen this analysis. Further exploration into the interaction between language models and CLIP’s visual encoder could also yield valuable insights. We see our analysis as a starting point.
\textbf{Second}, our study includes two academic training sources—WIT and LAION—for CLIP. Future work should investigate whether our findings generalize to other training sources, such as datasets generated by Stable Diffusion~\cite{rombach2021highresolution}, to advance our understanding of multi-modal dataset design.
\textbf{Lastly}, our analysis reveals a critical need for more refined fine-tuning strategies tailored to CLIP models, aimed at improving both classification accuracy and robustness.

\bibliographystyle{IEEEtran}
\bibliography{reference}

\end{document}